%% file: main.tex
\documentclass[10pt,twocolumn,letterpaper]{article}

\usepackage{main}
\usepackage{times}
\usepackage{epsfig}
\usepackage{graphicx}
\usepackage{url}
\usepackage{color}
\usepackage{amsfonts}
\usepackage{bm}
\usepackage{enumitem,amsthm,amsmath,amsfonts,amssymb}
\usepackage{mathtools}
\usepackage{xcolor}
\usepackage{subcaption}
\usepackage{makecell}
\usepackage[ruled, vlined, linesnumbered]{algorithm2e}%
\SetKwInOut{Parameter}{parameter}
\usepackage{caption}
\usepackage{subcaption}
\usepackage{multirow}
\usepackage{multicol}
\usepackage{rotating}
\usepackage{paralist}
\usepackage{comment}

\newtheorem{lemma}{Lemma}

\usepackage{diagbox}

\def\x{\mathbf{x}}
\def\y{\mathbf{y}}
\def\z{\mathbf{z}}
\def\ie{{\it i.e.}}
\def\eg{{\it e.g.}}
\def\asr{\texttt{ASR}}
\def\uasr{\texttt{UASR}}

\def\pert{\texttt{Pert}}

\let\emptyset\varnothing

\usepackage{url}
\usepackage{hyperref}

\iccvfinalcopy % *** Uncomment this line for the final submission

\def\iccvPaperID{6337} % *** Enter the ICCV Paper ID here

\ificcvfinal\fi

\begin{document}

%%%%%%%%% TITLE
%\title{Adversarial Attacks For Top-$k$ Multi-Label Learning}
\title{T$_k$ML-AP: Adversarial Attacks to Top-$k$ Multi-Label Learning}

\author{Shu Hu$^{1}$, \ \ \ Lipeng Ke$^{1}$, \ \ \  Xin Wang$^{2}$, \ \ \ Siwei Lyu$^{1}$\\
$^{1}$University at Buffalo, State University of New York \ \ \  $^{2}$Keya Medical\\
{\tt\small \{shuhu, lipengke, siweilyu\}@buffalo.edu}, \ \ \  {\tt\small xinw@keyamedna.com}
}

% \author{Shu Hu$^{1}$, Lipeng Ke$^{1}$, Xin Wang$^{2}$, and Siwei Lyu$^{1}$\\
% % \address{$^{1}$University at Buffalo, State University of New York, USA, {\tt\small \{shuhu,siweilyu\}@buffalo.edu}}
% $^{1}$University at Buffalo, SUNY, USA, {\tt\small \{shuhu,lipengke, siweilyu\}@buffalo.edu}\\
% $^{2}$Keya Medical, USA, {\tt\small xinw@keyamedna.com}
% % \and
% % Second Author\\
% % First line of institution2 address\\
% % {\tt\small secondauthor@i2.org}
% }

% \name{Shu Hu$^{1}$, Yuezun Li$^{2}$, and Siwei Lyu$^{1}$}
% \address{$^{1}$University at Buffalo, State University of New York, USA, {\tt\{shuhu,siweilyu\}@buffalo.edu} \\
% $^{2}$Ocean University of China, China, {\tt liyuezun@ouc.edu.cn}}

\maketitle
\ificcvfinal\thispagestyle{empty}\fi

%%%%%%%%% ABSTRACT
\begin{abstract}
   Top-$k$ multi-label learning, which returns the top-$k$ predicted labels from an input, has many practical applications such as image annotation, document analysis, and web search engine. However, the vulnerabilities of such algorithms with regards to dedicated adversarial perturbation attacks have not been extensively studied previously. In this work, we develop methods to create adversarial perturbations that can be used to attack top-$k$ multi-label learning-based image annotation systems (T$_k$ML-AP). Our methods explicitly consider the top-$k$ ranking relation and are based on novel loss functions. Experimental evaluations on large-scale benchmark datasets including PASCAL VOC and MS COCO demonstrate the effectiveness of our methods in reducing the performance of state-of-the-art top-$k$ multi-label learning methods, under both untargeted and targeted attacks. 
\end{abstract}

%%%%%%%%% BODY TEXT
\section{Introduction}

\begin{figure}[t]
\vspace{-\intextsep}
\centering
\includegraphics[trim=1 1 1 1, clip,keepaspectratio, width=0.45\textwidth]{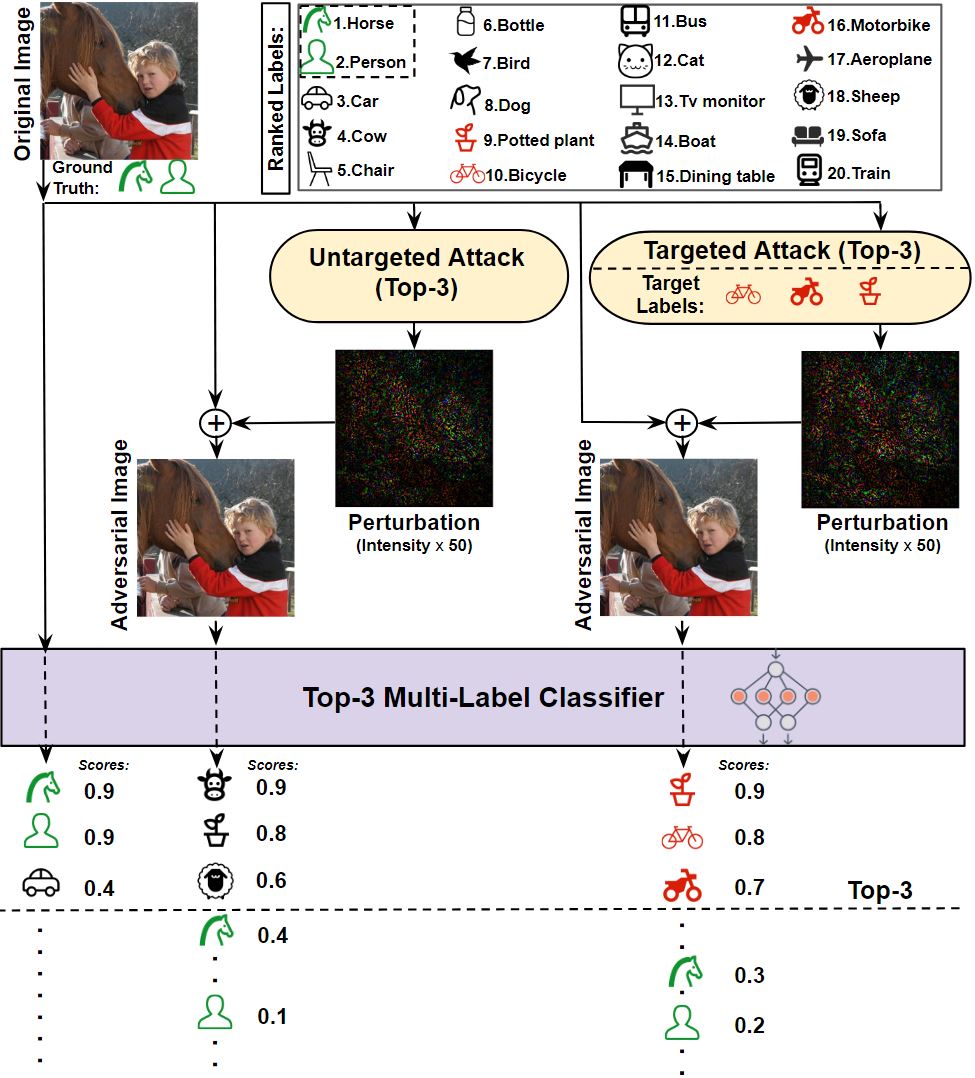}
\vspace{-1em}
\caption{\small \it Illustrative examples of the untargeted and targeted attacks to the top-$3$ multi-label image annotation for an image from the PASCAL VOC 2012 dataset. The green icons correspond to the ground truth labels. The red icons represent the targeted labels for attacking. The figure is better viewed in color.}
\vspace{-2em}
\label{fig:interpretation}
\end{figure}

The past decade has witnessed the {\it tour de force} of modern deep neural networks (DNNs), which have significantly improved, or in some cases, revolutionized, the state-of-the-art performance of many computer vision problems. Notwithstanding this tremendous success, the omnipotent DNN models are surprisingly vulnerable to adversarial attacks \cite{szegedy2013intriguing,goodfellow2014explaining,liu2016delving}. In particular, inputs with specially designed perturbations, commonly known as {\it adversarial examples}, can easily mislead a DNN model to make erroneous predictions.  The vulnerabilities of DNN models to adversarial examples impede the safe adoptions of machine learning systems in practical applications. It also motivates the explorations of algorithms generating adversarial examples \cite{carlini2017towards,moosavi2016deepfool,madry2017towards} as a means to analyze the vulnerabilities of DNN models and improve their security. 

Most existing works on generating adversarial examples have been focused on the case of multi-class classification \cite{akhtar2018threat, szegedy2013intriguing, goodfellow2014explaining,papernot2016limitations, carlini2017towards, moosavi2016deepfool}, where one instance can only be assigned to exactly one out of a set of mutually exclusive classes (labels). Because of the singleness of the labels, existing adversarial perturbation generation schemes for multi-class classification are based on the top-1 attack (\ie, C\&W \cite{carlini2017towards}, Deepfool \cite{moosavi2016deepfool}), only aiming to alter the top predicted label using the adversarial perturbation. 

However, in many real-world applications such as image annotation, document categorization, and web search engines, it is more natural to solve the multi-label learning problem, where an instance is associated with a non-empty subset of labels. Furthermore, in these applications, the output of the system is usually a set of labels of a fixed size, corresponding to the top-$k$ predicted labels. We term this as the {\it top-$k$ multi-label learning} (T$_k$ML). The practical cases of T$_k$ML open more opportunities for attackers and leading to larger uncertainties for defenders. There are two common settings that we will consider subsequently for T$_k$ML adversarial attacks. The {\it untargeted attack} aims to only replace the top-$k$ labels with a set of arbitrary $k$ labels that are not true classes of the untampered input. The {\it targeted attack}, on the other hand, aims to coerce the T$_k$ML classifier to use a specific set of $k$ labels that are not true classes of the input as the top-$k$ predictions.

In this work, we describe the first untargeted and targeted adversarial attack algorithms for T$_k$ML based on a continuous formulation of the ranking operation, which we term as T$_k$ML-AP. Specifically, we note that to perturb the predictions of a T$_k$ML algorithm, it is sufficient to clear any ground-truth labels from the top-$k$ set. There are many different ways to achieve this, but we will focus on ones that enlist the ``least actions’’, \ie, perturbing the predicted labels with minimum changes to the original label rankings. For the untargeted attack, this means move the ground-truth labels out of the top-$k$ predictions, and for the targeted attack, this means move the target labels to the top-$k$ set. Fig.\ref{fig:interpretation} gives an illustrative explanation of the proposed idea. 

Thus, the key challenge in generating adversarial examples for T$_k$ML is to optimize perturbations that can lead to the change of top-$k$ rankings of the predicted label. To this end, we introduce a reformulation of the top-$k$ sum that lends itself to efficient numerical algorithms based on gradient descent methods. In particular, we provide loss functions for adversarial perturbations to T$_k$ML that are convex in terms of the individual prediction scores.  This has a further advantage that even though the model may be nonlinear, a convex loss function can encourage many equally effective local optima. Hence any adversarial perturbation that can lead the model to have the same loss value will have equal effects. We demonstrate the effectiveness of our method on attacking state-of-the-art T$_k$ML algorithms using large scale benchmark datasets (PASCAL VOC 2012 \cite{everingham2015pascal} and MS COCO 2014 \cite{lin2014microsoft}). The main contributions of our work can be summarized as follows: 
\begin{compactenum}
\item We present the first algorithms for untargeted and targeted adversarial attacks to the T$_k$ML problem.
\item Our method is based on a continuous reformulation of the non-differentiable ranking operation. The objective function is convex in terms of the individual prediction scores, which is easier to optimize.
\item Numerical experiments on large-scale benchmark datasets confirm the effectiveness of our method in attacking state-of-the-art T$_k$ML algorithms. 
\end{compactenum}

\section{Backgrounds}
\subsection{Top-$k$ Multi-label Learning (T$_k$ML)}
\label{sec:T$_k$ML}

Let us assume a general multi-label classification problem with a total of $m > 1$ possible labels. For an input $\x \in \mathbb{R}^d$, its true labels are represented by a binary label vector $\y = [y_1,y_2,\cdots,y_m]^\top \in \{0, 1\}^m$, with $y_j=1$ indicating that $\x$ is associated with the $j$-th label. We also use $Y=\{j|y_j=1\}$ to represent the set of true labels of $\x$. Note that $Y$ and $\y$ are equivalent notations of the truth labels of $\x$.

We introduce a continuous multi-label prediction function $F(\x) = [f_1(\x), f_2(\x), \cdots, f_m(\x)]^\top$, with each $f_j(\x) \in [0,1]$ corresponding to the prediction score of $\x$ with regards to the $j$-th class\footnote{Here we assume the prediction scores are {\it calibrated}, \ie, taking values in the range of $[0,1]$. For $f \in \mathbb{R}$, we can use simple transforms such as ${1 \over 1+e^{-f}}$ to map it to the range of $[0,1]$ without changing their ranking.}. We denote $[f_{[1]}(\x), f_{[2]}(\x), \cdots, f_{[m]}(\x)]^\top$ as the sorted values of $F(\x)$ in descending order, \ie, $f_{[1]}(\x)$ is the largest (top-$1$) score, $f_{[2]}(\x)$ is the second largest (top-$2$) score, and so on. Furthermore, $[j]$ corresponds to the label index of the top-$j$ prediction score, \ie, $j' = [j]$ if $f_{j'}(\x) = f_{[j]}(\x)$. In ranking the values, ties can be broken in any consistent way. For input $\x$, the top-$k$ multi-label classifier returns the set $\hat{Y}_k(\x) = \{[1], \cdots, [k]\}$ for $1 \le k < m$. In other words, we can convert a general multi-label predictor $F$ to a top-$k$ multi-label classifier by returning the set of labels corresponding to the set of top-$k$ prediction scores from $F(\x)$. {This problem is related to many types of learning problems. If $|Y|=1$, $k=1$, it becomes the conventional multi-class problem. If $|Y|=1$, $k\geq 1$, it becomes top-k multi-class problem \cite{lapin2015top}.  If  $k=|Y|$, $|Y|\geq 1$, it becomes the conventional multi-label problem.} The top-$k$ setting is often implicitly used in applications of multi-label learning. For instance, in image annotation, when the number of possible labels is large, the system often returns a fixed number of top annotations that are most relevant to the image\footnote{Another strategy in multi-label learning is to return all labels with prediction score above a preset threshold. The result will be a list of labels of varying length. The top-$k$ multi-label classification can be regarded as using a varying threshold to fix the number of the returned labels.}. 

A successful top-$k$ multi-label classification should lead to consistency between the true labels ($Y$) and the predicted labels $\hat{Y}_k(\x)$ of the input. The situation is complicated by the difference in size of $Y$ and $\hat{Y}_k(\x)$, so we use the following criterion: when $k \ge |Y|$, it corresponds to $Y \subseteq \hat{Y}_k(\x)$; when $k \le |Y|$, it is the case $\hat{Y}_k(\x) \subseteq Y$. In other words, one is the subset of the other depending on the relation of $k$ and the number of the truth labels. We define the top-$k$ label consistency score as:
\begin{equation}
E(Y,\hat{Y}_k(\x)) =\mathbb{I}_{Y \subset \hat{Y}_k(\x)} +  \mathbb{I}_{\hat{Y}_k(\x) \subset Y} +  \mathbb{I}_{Y = \hat{Y}_k(\x)},
\label{e:E}
\end{equation}
where $\mathbb{I}_{c}$ is the indicator function that takes value $1$ when condition $c$ is true and $0$ otherwise. As such, $E(Y,\hat{Y}_k(\x))$ is $1$ for a successful multi-label classification of input $\x$ and $0$ otherwise.

% This formulation is related to previous adversarial attack methods that are widely used to generate adversarial perturbations.  
% \begin{compactitem}
% \item If $|Y|=1$, $k=1$, and transform the constraint as a penalty of the objective function, then it becomes the classical C\&W method \cite{carlini2017towards}.
% \item If $|Y|=1$, $k>1$, transform the constraint as a penalty of the objective function \shu{and restrict the order of top-$k$ attack labels}, our problem reduce to the problems of CW$^{k}$ \cite{zhang2020learning}.
% \item If $|Y|=1$, $k>1$, and without consider the constraint of the upper bound of $\|\z\|_2$ , this is the $k$Fool \cite{tursynbek2020geometry} problem.
% \item If $k=m-|Y|$, $|Y|\geq 1$, our problem reduce to the extreme case of the problem in ML-AP \cite{song2018multi}, which attempts to change all labels. 
% %In other words, it takes more actions to push the ground truth labels to the end of the sorted descending prediction label list.
% \end{compactitem}

\subsection{Top-$k$ and Average Top-$k$} \label{sec:topk}

Top-$k$ ranking emerges as a natural element in the learning objectives in various problems such as multi-class learning and robust binary classification \cite{lapin2015top,lyu2018univariate,hu2020learning}. However, as a function of all elements in a set, the top-$k$ ranking function is non-continuous, non-differentiable, and non-convex. This makes the optimization involving top-$k$ ranking challenging. 

To mitigate these problems of the top-$k$ operator, we can use the {\it average of top-$k$} function \cite{fan2017learning}, which is defined for a set $F = \{f_1,\cdots,f_m\}$ as 
\begin{equation} \textstyle
\phi_k(F) = {1 \over k} \sum_{j=1}^k f_{[j]}.
    \label{e:atk}
\end{equation}
It is not difficult to show that (i) $\phi_k(F) \ge f_{[k]}$, and (ii) $\phi_k(F) = f_{[k]}$ when $f_{[1]} = \cdots = f_{[k]}$. As such, the average of top-$k$ is a tight upper-bound of the top-$k$. It can be proved that $\phi_k(F)$ is a convex function of the elements of $F$ \cite{boyd2004convex}. More importantly, it affords an equivalent form as an optimization problem \cite{ogryczak2003minimizing}.
\begin{lemma} 
For $f_i(\x) \in [0,1]$, we have
\begin{eqnarray}
\phi_k(F) &= {1 \over k} \min_{\lambda \in [0,1]}\{k\lambda+\sum_{j=1}^m[f_j-\lambda]_+\} \label{e:atk}\\
f_{[k]} & \in \arg\!\min_{\lambda \in [0,1]}\{k\lambda+\sum_{j=1}^m[f_j-\lambda]_+\}, 
\label{e:ftk}
\end{eqnarray}
where $[a]_+ = \max\{0,a\}$ is the hinge function.
\label{lemma:convex}
\end{lemma}
For completeness, we include the proof of Lemma \ref{lemma:convex} in Appendix \ref{append:proof_lemma_1}.  Lemma \ref{lemma:convex} enables us to incorporate the average top-$k$ function in conventional sub-gradient based optimization. 

\subsection{Related Works}

Due to the limit of space, we only provide a brief overview of relevant works. A full survey of adversarial attacks on deep learning models can be found in \cite{akhtar2018threat}. The major differences between the work in this paper and the related works are summarized in Table \ref{tab:comparison}.

Most existing adversarial attacking methods target multi-class classification problems (corresponding to the special case of T$_k$ML with $k=1$ and $|Y| = 1$ for all inputs). As such, these methods often target the top prediction and aim to change it with perturbations. For untargeted attacks, DeepFool \cite{moosavi2016deepfool} is a generalization of the minimum attack under general decision boundaries by
%. This method sometimes just
swapping labels from the top-$2$ prediction. The work of \cite{moosavi2017universal} (UAP) aims to find universal adversarial perturbations that are independent of individual input images. Both DeepFool and UAPs are top-1 multi-class adversarial attack methods. For targeted attacks, FGSM \cite{goodfellow2014explaining} and I-FGSM \cite{kurakin2016adversarial} are two popular attack schemes that use the gradient of the DNN models with regards to the input to generate adversarial samples. The CW method \cite{carlini2017towards} improves on the previous method by using regularization and modified constraints.

Realizing that only attacking the top predictions may not be effective, several works introduce attacks to the top-$k$ (for $k > 1$) predictions in a multi-class classification system. 
$k$Fool \cite{tursynbek2020geometry} and CW$^k$ \cite{zhang2020learning} extend the original DeepFool \cite{moosavi2016deepfool} and CW \cite{carlini2017towards} methods to exclude the truth label out of the top-$k$ predictions. $k$Fool is based on a geometry view of the decision boundary between $k$ labels and the truth label in the multi-class problem. The UAP method is extended in \cite{tursynbek2020geometry} to top-$k$ Universal Adversarial Perturbations ($k$UAPs). In addition, the CW method is extended to a top-$k$ version known as CW$^k$ in \cite{zhang2020learning}. However, all these methods are still designed for multi-class classification (\ie $|Y| = 1$), and cannot be directly adapted to the attacks to the more general top-$k$ multi-label learning.

The authors of \cite{song2018multi} describes an adversarial attack to multi-label classification extending existing attacks to multi-class classification. This method is further studied in \cite{zhou2020generating}, which transfers the problem of generating attack to a linear programming problem.  To make the predictions of adversarial examples lying inside of the training data distribution, \cite{melacci2020can} proposed a multi-label attack procedure with an additional domain knowledge-constrained classifier. These are all for multi-label learning without the top-$k$ constraint. Our experiments in Section \ref{sec:exp} show that they are not effective for the top-$k$ setting. 

%Although the proposed model applies to multi-label learning, it cannot change all positive labels in the ground truth label set while adding a perturbation.

\begin{comment}
In extreme multi-label classification, due to feature sparsity problem, the features of the instances in the training set  are quite different from those in the test set.   \cite{babbar2018adversarial} adopted multi-label adversarial examples for adversarial training to improve the robustness of the models. 

Similarly, \cite{wu2017adversarial} applies  adversarial examples for adversarial training in multi-label learning framework. They try to improve the robustness of natural language processing models with worst-case perturbations and also do not change the predicted labels in the training data.

 \cite{rao2020thorough} applies the targeted attack methods from ML-AP\cite{song2018multi} on medical images on the multi-label classification task.
\end{comment}

%However, the goal of our methods are trying to change the training labels with small perturbations on the data.

\begin{table}[t]
\scriptsize
% \captionsetup{font=footnotesize}
\centering
\setlength\tabcolsep{3.5pt}
% \scriptsize{
\begin{tabular}{|c|ccccc|}
\hline
\diagbox{Methods}{Features} & \makecell{Multi \\ Label}  & \makecell{Untargeted \\ Attack}& \makecell{Universal \\ Attack}& \makecell{Targeted \\ Attack} & Top-$k$ \\ \hline
$k$Fool \cite{tursynbek2020geometry} & $\times$ & $\checkmark$ &$\times$ & $\times$& $\checkmark$ \\ 
$k$UAPs \cite{tursynbek2020geometry} & $\times$ & $\checkmark$ &$\checkmark$ & $\times$& $\checkmark$ \\ 
CW$^{k}$ \cite{zhang2020learning} & $\times$ & $\times$ &$\times$&  $\checkmark$& $\checkmark$ \\ 
 ML-AP \cite{song2018multi}& $\checkmark$ & $\times$& $\times$ &$\checkmark$ & $\times$ \\ \hline
 \textbf{T$_k$ML-AP-U} (this paper)& $\checkmark$ &  $\checkmark$ &$\times$ &$\times$ & $\checkmark$ \\ 
 \textbf{T$_k$ML-AP-Uv} (this paper)& $\checkmark$ &  $\checkmark$ &$\checkmark$ &$\times$ & $\checkmark$ \\ 
 \textbf{T$_k$ML-AP-T} (this paper)& $\checkmark$ &  $\times$ &$\times$  &$\checkmark$& $\checkmark$ \\ \hline
\end{tabular}
\vspace{-1em}
\caption{\it \small Summary of the difference between previous works with our methods (T$_k$ML-AP).}
\label{tab:comparison}
% }
\vspace{-2em}
\end{table}

\section{Method}

In this work, we introduce new methods to generate adversarial perturbations to attack top-$k$ multi-label classification. We term our method as T$_k$ML-AP. Unlike the multi-label adversarial perturbation method in \cite{song2018multi}, we consider the top-$k$ ranking in the T$_k$ML problem an essential requirement to design loss functions in our methods. Hence, in our methods, the ranking relation is explicitly handled, using the results in Section \ref{sec:topk}. Specifically, we describe our methods in detail for the instance-specific and instance-independent (universal) untargeted attacks (T$_k$ML-AP-U and T$_k$ML-AP-Uv) and targeted attacks (T$_k$ML-AP-T). A comparison with previous works is given in Table \ref{tab:comparison}.

\subsection{Untargeted Attack} 
\label{sec:UA}

\noindent{\bf Formulation}. The untargeted attack to top-$k$ multi-label learning (T$_k$ML-AP-U) aims to find a minimum perturbation to the input that can push the prediction scores of the truth labels outside of the top-$k$ set. It can be formulated as finding a perturbation signal $\z$ for an input $\x$ such that 
\begin{equation}
    \min_\z \|\z\|_2, ~\text{s.t.}~ E(Y,\hat{Y}_k(\x+\z)) = 0,
    \label{e:E1}
\end{equation}
where $E(Y,\hat{Y}_k)$ is defined in Eq.\eqref{e:E}\footnote{Note that the definition is minimal: it only changes labels that are correctly predicted by $F(\x)$, \ie $Y \cap \hat{Y}_k(\x)$. True labels that are incorrectly predicted by $F$ and not in $\hat{Y}_k(\x)$ are expected to be intact.}. Because $k\le m-1$, we can rewrite Eq.\eqref{e:E1} with a more revealing equivalent form,  
\begin{equation}
    \min_\z {1 \over 2}\|\z\|_2^2, ~\text{s.t.}~ \max_{j \in Y}f_j(\x+\z) \le f_{[k+1]}(\x+\z).
    \label{e:E2}
\end{equation}
Note that the constraint is equivalent to have $Y\subseteq \{j|f_j(\x+\z)< f_{[k]}(\x+\z)\}$, the converse of which means at least one truth label is inside the top-$k$ range. 

\noindent{\bf Relaxation}. The optimization problem in Eq.\eqref{e:E2} is difficult to optimize directly, so we introduce the top-$k$ multi-label loss, $\left[\max_{j \in Y}f_j(\x+\z) - f_{[k+1]}(\x+\z)\right]_+$, as a surrogate to the constraint. The top-$k$ multi-label loss precisely reflects the requirement to exclude the true labels out of the top-$k$ range. It is zero when the maximal prediction score from the true labels is no greater than the $k+1$-th prediction score, and positive otherwise. Rewriting the objective function using the Lagrangian form, we have
%In addition, as in \cite{su2019one,sabour2015adversarial}, we use a more convenient and equivalent form that minimize the top-$k$ multi-label loss 
\begin{equation}
    %\min_\z \left[\max_{j \in Y}f_j(\x+\z) - f_{[k+1]}(\x+\z)\right]_+, ~\text{s.t.}~ \|\z\|_2 \le \epsilon,
    \min_\z {\beta \over 2}\|\z\|_2^2+ \left[\max_{j \in Y}f_j(\x+\z) - f_{[k+1]}(\x+\z)\right]_+, 
\label{eq:AP-obj}
\end{equation}
where $\beta > 0$ is a prechosen trade-off parameter. 

\smallskip
\noindent{\bf Optimization}. 
The ranking operation in Eq.(\ref{eq:AP-obj}) can be further removed. Specifically, denote $t_{m+1-j}=[\max_{y\in Y}f_y(\x+\z)-f_{j}(\x+\z)]_+$ for $j=1,\cdots,m$. With a bit of abuse of the notation, we denote $t_{[j]}$ as the top-$j$ element in the set $\{t_1,\cdots,t_m\}$\footnote{Note that $[j]$ in $t_{[j]}$ may not correspond to the same index as in the case of $f_{[j]}$ as it depends on the ranking of different sets.}. Note that there is a simple correspondence, as $t_{[m-k]} = [\max_{y\in Y}f_y(\x+\z)-f_{[k+1]}(\x+\z)]_+$. 

As shown in Section \ref{sec:topk}, we have the following bound of the top-$k$ value using the average of $\{t_1,\cdots,t_m\}$, as ${1 \over m-k}\sum_{j=1}^{m-k} t_{[j]} \ge t_{[m-k]} = [\max_{y\in Y}f_y(\x+\z)-f_{[k+1]}(\x+\z)]_+$. Furthermore, using Lemma \ref{lemma:convex}, we can rewrite the average of top-$(m-k)$ elements of $\{t_1,\cdots,t_m\}$ as ${1 \over m-k}\sum_{j=1}^{m-k} t_{[j]} = \min_{\lambda \in [0,1]} \lambda+{1 \over m-k}\sum_{j=1}^m[t_{j}-\lambda]_+$. Replacing the definition of $t_j$, the inner term $[[\max_{y\in Y}f_y(\x+\z)-f_{[m+1-j]}(\x+\z)]_+-\lambda ]_+$ can be further simplified by removing the double hinge function to
$[\max_{y\in Y}f_y(\x+\z)-f_{[j]}(\x+\z)-\lambda]_+$, according to the following result. 
\begin{lemma} 
For $\lambda \ge 0$, $[[a-x]_+-\lambda]_+ = [a-x-\lambda]_+$.
\label{lemma:2}
\end{lemma}
The proof of Lemma \ref{lemma:2} is deferred to Appendix \ref{append:proof_lemma_2}. Putting all results together, we get the objective function for finding adversarial perturbation in an untargeted attack to the top-$k$ multi-label learning as:
\begin{equation}\small
\min_{\lambda \in [0,1],\z} {\beta \over 2}\|\z\|_2^2 +  \lambda+{1 \over m-k}\sum_{j=1}^m[\max_{y\in Y}f_y(\x+\z)-f_{j}(\x+\z)-\lambda]_+.
\label{e:E3}
\end{equation}

This optimization problem can be solved with an iterative gradient descent approach \cite{hu2020learning, fan2017learning}. We initialize $\z$ and $\lambda$, then update them with the following steps: 
\begin{equation}\small
\begin{aligned}
\z_{l+1} = &(1-\beta\eta_l)\z_l- \frac{\eta_l}{m-k}\sum_{j=1}^m \left(\frac{\partial f_{y'}(\x^\prime)}{\partial \x^\prime}-\frac{\partial f_{j}(\x^\prime)}{\partial \x^\prime}\right)\Bigg|_{\x^\prime=\x+\z_l}\\ &\cdot\mathbb{I}_{[f_{y'}(\x+\z_l)-f_{j}(\x+\z_l)>\lambda_l]},\\
\lambda_{l+1} =& \lambda_l - \eta_l\cdot \Big(1-\frac{1}{m-k}\sum_{j=1}^m \mathbb{I}_{[f_{y'}(\x+\z_l)-f_{j}(\x+\z_l)>\lambda_l]}\Big),
\end{aligned}
\label{eq:update_rules}
\end{equation}
where $\eta_l$ is the step size and $y' \in \max_{y\in Y}$. 
This iterative process continues until the termination conditions are met. The overall procedure is described in detail in Algorithm \ref{Alg1}.

% This optimization problem can be solved with a projected gradient descent (PGD) approach. First, we use the gradient descent on the second and the third terms to update the parameters $\z$ and $\lambda$ with the following steps:
% \begin{equation}\small
% \begin{aligned}
% \z_{l+1} = &(1-\beta)\z_l- \frac{\eta_l}{m-k}\sum_{j=1}^m \left(\frac{\partial f_{y'}(\x^\prime)}{\partial \x^\prime}-\frac{\partial f_{j}(\x^\prime)}{\partial \x^\prime}\right)\Bigg|_{\x^\prime=\x+\z_l}\cdot\\ &\mathbb{I}_{[f_{y'}(\x+\z_l)-f_{j}(\x+\z_l)>\lambda_l]},\\
% \lambda_{l+1} =& \lambda_l - \eta_l\cdot \Big(1-\frac{1}{m-k}\sum_{j=1}^m \mathbb{I}_{[f_{y'}(\x+\z_l)-f_{j}(\x+\z_l)>\lambda_l]}\Big),
% \end{aligned}
% \label{eq:update_rules}
% \end{equation}
% where $\eta_l$ is the step size and $y' \in \max_{y\in Y}$. Second, we apply projection on the $\z_{l+1}$ with a projector operation $\mathcal{P}_{\epsilon}$ controls the criteria $\|\z_{l+1}\|_2\leq\epsilon$, where $\epsilon$ is a predefined threshold. This process iteratively continues until the termination conditions are met. The overall procedure is described in detail in Algorithm \ref{Alg1}.

\begin{algorithm}[t]
    \caption{Untargeted Attack (T$_k$ML-AP-U)}\label{Alg1}
    \SetAlgoLined
    % \begin{flushleft}
    \KwIn{$\x$, predictor $F$, $k$, $\eta_l$, $\beta$}
    \KwOut{adversarial example $\x^*$, perturbation $\z^*$} 
    % \textbf{Input:} $\textbf{x}$, predictor $F$, set $P$, maxiter   \\
    % \textbf{Output:} adversarial example $\textbf{x}^*$, perturbation $\z^*$  \\
    \textbf{Initialization:} $l=0$, $\x^* = \x$, $\z_0$, and $\lambda_0$ \\

    % \end{flushleft}
    
    % \While{$l\leq $ maxiter}{
    \While {$E(Y,\hat{Y}_k(\x+\z)) \neq 0$}{
    Compute $\z_{l+1}$ and $\lambda_{l+1}$ with Eq.(\ref{eq:update_rules});
    % $\z_{l+1} = \z_l-\eta_l\sum_{j=1}^m \left.\frac{\partial (\max_{y\in Y}f_y(\x^\prime)-f_{j}(\x^\prime))}{\partial \x^\prime}\right|_{\x^\prime=\x+\z_l}\cdot \mathbb{I}_{[\max_{y\in Y}f_y(\x+\z_l)-f_{j}(\x+\z_l)>\lambda_l]}$;

    % $\lambda_{l+1} = \lambda_l - \eta_l\cdot \Big(m-k-\sum_{j=1}^m \mathbb{I}_{[\max_{y\in Y}f_y(\x+\z_l)-f_{j}(\x+\z_l)>\lambda_l]}\Big)$;
    
    % \If{Proj==True}{
    
    % $\z_{l+1} = \mathcal{P}_{\epsilon}(\z_{l+1})$;
    % }
    
    $\x^* = \x+\z_{l+1}$, $\z^* = \z_{l+1}$;
    
    $l = l+1$;
    
    }
    \Return{$\x^*$, $\z^*$}
\end{algorithm}

\noindent\textbf{Universal untargeted attack}. We can extend the instance-specific untargeted attack to a universal adversarial attack that is independent of the input \cite{moosavi2017universal} so can be shared by all instances. Specifically, given a dataset $\textbf{X}=\{\x_1,\cdots,\x_n\}$ and its ground truth label set $Y=\{Y_1,\cdots,Y_n\}$, where $Y_i:=Y(\x_i)$, finding the instance-independent (universal) adversarial perturbation $\z$ is formulated as $\min_{\lambda \in [0,1],\z} {1 \over n} \sum_{i=1}^n L(\z,\lambda; \x_i,Y_i)$, where $L(\z,\lambda; \x_i,Y_i)$ is the objective function in Eq.\eqref{e:E3}. The solution to the universal untargted attack can be obtained using a similar procedure based on Algorithm \ref{Alg1}. Please refer to Appendix \ref{append:UUA_algorithm} for the details about the T$_k$ML-AP-Uv algorithm.

\subsection{Targeted Attack}
\label{targeted_attack}

\noindent\textbf{Formulation.} We next consider the targeted attack, the aim of which is to plant a set of $k$ labels, $\widetilde{Y} \subset \{1,\cdots,m\}$ and $\widetilde{Y} \cap Y = \emptyset$, as the top-$k$ predictions. We formulate the learning objective of the targeted attack on top-$k$ multi-label learning (T$_k$ML-AP-T) as 
\begin{equation}
    \min_\z \|\z\|_2, ~\text{s.t.}~ \widetilde{Y} = \hat{Y}_k(\x+\z).
\label{eq:ta}
\end{equation}
The constraint in Eq.(\ref{eq:ta}) exactly reflects the requirement that the top-$k$ predicted labels of the perturbed are all from the targeted label set. 

\smallskip
\noindent\textbf{Relaxation.} Analogous to the untargeted attack, we rewrite the objective function into a form that lends itself to optimization.  Specifically, if we have $\widetilde{Y} = \hat{Y}_k(\x+\z)$, it means that the prediction scores of labels in $\widetilde{Y}$ occupy the top-$k$ ranks. So the sum of prediction scores from labels in the sets $\widetilde{Y}$ and $\hat{Y}_k(\x+\z)$ are also the same, \ie, we have $\sum_{j=1}^{k} f_{[j]}(\x+\z)-\sum_{j\in \widetilde{Y}} f_{j}(\x+\z) = 0$. Furthermore, if $\widetilde{Y} \not= \hat{Y}_k(\x+\z)$, $\sum_{j=1}^{k} f_{[j]}(\x+\z)-\sum_{j\in \widetilde{Y}} f_{j}(\x+\z) \ge 0$ as by definition, the second term cannot be greater than the first term. This suggest that $\sum_{j=1}^{k} f_{[j]}(\x+\z)-\sum_{j\in \widetilde{Y}} f_{j}(\x+\z)$ is a surrogate to the constraint in Eq.(\ref{eq:ta}). It is zero when all target attacked labels are in the top-$k$ positions, and positive otherwise. Introducing the Lagrangian form, we can reformulate Eq.(\ref{eq:ta}) as
\begin{equation} \textstyle
    \min_\z {\beta \over 2} \|\z\|_2^2 + \sum_{j=1}^{k} f_{[j]}(\x+\z)-\sum_{j\in \widetilde{Y}}f_{j}(\x+\z),
    \label{E:E5}
\end{equation}
where $\beta > 0$ is a prechosen trade-off parameter. Note that the second term in Eq.\eqref{E:E5} is precisely the sum of top-$k$ elements. Using the results of Lemma \ref{lemma:convex}, we can remove the explicit ranking operation in Eq.\eqref{E:E5}. Specifically, we have $\sum_{j=1}^{k} f_{[j]}(\x+\z) = \min_{\lambda\in[0,1]}\{k\lambda +\sum_{j=1}^m[f_j(\x+\z)-\lambda]_+\}$. Further simplifying the last two terms in Eq.\eqref{E:E5} yields
\[
\begin{aligned} 
     & \textstyle \Big\{k\lambda +\sum_{j=1}^m[f_j(\x+\z)-\lambda]_+\Big\}-\sum_{j\in\widetilde{Y}}f_{j}(\x+\z)\\
     =& \textstyle \sum_{j\in\widetilde{Y}}\Big([f_j(\x+\z)-\lambda]_+-(f_j(\x+\z)-\lambda)\Big)\\
     +&\textstyle \sum_{j\not\in \widetilde{Y}}[f_j(\x+\z)-\lambda]_+\\
     =&\textstyle\sum_{j\in\widetilde{Y}}[\lambda - f_j(\x+\z)]_+ + \sum_{j\not\in \widetilde{Y}}[f_j(\x+\z)-\lambda]_+,
\end{aligned}
\]
where we use a fact that $[a]_+-a=[-a]_+$. Introducing $s_j = 2\mathbb{I}_{j \in \widetilde{Y}} - 1 \in\{-1,1\}$, we can rewrite Eq.\eqref{E:E5} more concisely as
\begin{equation} \textstyle
     \min_{\lambda\in[0,1],\z} {\beta \over 2}\|\z\|_2^2+\sum_{i=1}^m[s_j(\lambda-f_j(\x+\z))]_+
\label{eq:UA_final}
\end{equation}

%We introduce the top-k multi-label target attacked loss as a surrogate to this constraint, which is defined as $\sum_{j\in\widetilde{Y}}\sum_{\overline{j}\in N} \mathbb{I}_{[f_j(\x+\z)<f_{\overline{j}}(\x+\z)]}$. This loss intuitively reflects the ground truth labels are not included in the top-$k$ range. It is zero when all scores in $N$ are smaller than scores in $P$, and positive otherwise. However, to calculate it, we need to do pairwise comparisons of scores in $P$ and scores in $N$, which lead the optimization algorithms with quadratic running time complexity. Therefore, we use another surrogate ranking loss

\smallskip
\noindent\textbf{Optimization.} This optimization problem can also be solved with an iterative gradient descent approach as in the untargeted attack. We initialize $\z$ and $\lambda$, then update them with the following steps: 
\begin{equation}\small
    \begin{aligned}
    \z_{l+1} = &(1-\beta\eta_l)\z_l\\
    &-\eta_l\sum_{j=1}^m (-s_j) \left.\frac{\partial f_j(\x^\prime)}{\partial \x^\prime}\right|_{\x^\prime = \x+\z_l}\cdot
    \mathbb{I}_{[s_j(\lambda_l-f_j(\x+\z_l))>0]}\\
    \lambda_{l+1} = &\lambda_l - \eta_l\sum_{j=1}^m s_j\cdot \mathbb{I}_{[s_j(\lambda_l-f_j(\x+\z_l))>0]}
    \end{aligned}
\label{eq:TA_update_rules}
\end{equation}
where $\eta_l$ is the step size. The overall procedure is described in detail in Algorithm \ref{Alg0}. The algorithm stops when the termination conditions are met.

\begin{algorithm}[ht]
    \caption{Targeted Attack (T$_k$ML-AP-T)}\label{Alg0}
    \SetAlgoLined
    % \begin{flushleft}
    \KwIn{$\x$, predictor $F$, $\widetilde{Y}$, max\_iter, $\eta_l$, $\beta$}
    \KwOut{adversarial example $\x^*$, perturbation $\z^*$} 
    % \textbf{Input:} $\textbf{x}$, predictor $F$, set $P$, maxiter   \\
    % \textbf{Output:} adversarial example $\textbf{x}^*$, perturbation $\z^*$  \\
    \textbf{Initialization:} $l=0$, $\x^* = \x$, $\z_0$, and $\lambda_0$ \\

    % \end{flushleft}
    
    % \While{$l\leq $ maxiter}{
    \While {$l\leq $ max\_iter}{
    Compute $\z_{l+1}$ and $\lambda_{l+1}$ with Eq.(\ref{eq:TA_update_rules});
    % $\z_{l+1} = \z_l-\eta_l\Big(\sum_{j=1}^m z_j \left.\frac{\partial f_j(\x^\prime)}{\partial \x^\prime}\right|_{\x^\prime = \x+\z_l}\cdot \mathbb{I}_{[z_j(\lambda_l-f_j(\x+\z_l))>0]}\Big)$;

    % $\lambda_{l+1} = \lambda_l - \eta_l\sum_{j=1}^m z_j\cdot \mathbb{I}_{[z_j(\lambda_l-f_j(\x+\z_l))>0]}$;
    
    % $\z_{l+1} = \mathcal{P}_{\epsilon}(\z_{l+1})$;
    
    $\x^* = \x+\z_{l+1}$, $\z^* = \z_{l+1}$;
    
    $l = l+1$;
    
    }
    \Return{$\x^*$, $\z^*$}
\end{algorithm}

\section{Experiments}
\label{sec:exp}

We evaluate the performance of the proposed adversarial attacks (\ie, T$_k$ML-AP-U, T$_k$ML-AP-Uv, and T$_k$ML-AP-T) in the practical problem of image annotation, the goal of which is to predict the labels of an input image. 
% For the purpose of review, the source code is accessible in the supplementary file. 
Due to the limit of the space, we present the most significant information and results of our experiments, with more detailed information and additional results in the complementary materials\footnote{Code: \url{https://github.com/discovershu/TKML-AP}.}. 

\subsection{Experimental Settings}
\label{sec:exp-set}

\noindent \textbf{Datasets and baseline models.} Our experiments are based on two popular large-scale image annotation datasets, namely PASCAL VOC 2012 \cite{everingham2015pascal} and MS COCO 2014 \cite{lin2014microsoft}. Both datasets have multiple true labels associated with each image: the average number of positive labels per instance in PASCAL VOC 2012 and MS COCO 2014 are 1.43 (out of 20) and 3.67 (out of 80), respectively. All RGB images are with pixel intensities in the range of $\{0, 1, \cdots, 255\}$.

On the two datasets, we train deep neural network-based top-$k$ multi-label classifiers as baseline models. For the PASCAL VOC 2012 dataset, similar to \cite{song2018multi}, we adopt the inception-v3 \cite{szegedy2016rethinking} model pre-trained on ImageNet \cite{russakovsky2015imagenet} and fine-tuned on PASCAL VOC 2012. For MS COCO 2014 datasets, we use a ResNet50 \cite{he2016deep} based model. Both models are originally designed for multi-class classification, so we convert them to multi-label models by replacing the softmax layer with sigmoid classification layer as in \cite{ridnik2021tresnet}\footnote{After retraining the models, we get 0.934 mAP performance for PASCAL VOC 2012 and 0.867 mAP performance for  MS COCO 2014 on the corresponding validation datasets, which are close to the state-of-the-art performance \cite{song2018multi,ridnik2021tresnet}.}. We further modify the model to output the top-$k$ predicted labels. %More details about the settings can be found in Appendix \ref{append:baseline_methods_settings}.
%It is worth noted that we do not focus on some conventional multi-label learning models because they are label transformation-based methods, which could be separated as several uncorrelated multi-class classification components and easily attacked by traditional multi-class attacking methods as mentioned in \cite{song2018multi}. 

We select 1,000 images from the validation set in PASCAL VOC 2012 and MS COCO 2014 datasets respectively as the test set to test the untargeted and targeted attack methods. These images are correctly predicted by the baseline T$_k$ML models, \ie, the predicted top-$k$ labels either contain or are completely from the true labels. For the universal untargeted attack, however, we need a training dataset to find the universal perturbation. Therefore, we select 3,000 images from the validation set of MS COCO 2014 as the training set and evaluate the attack performance on another different 1,000 images from the same validation set.  For the targeted attacks, we choose the target labels as in \cite{carlini2017towards, zhang2020learning}, where we consider three different strategies (see Fig.\ref{fig:TA} for more details).
\begin{compactitem}
\item Best Case. In this case, we select $k$  labels that are not true labels and have the highest prediction scores. These labels are the runner-ups and the regarded as the easiest labels to attack. 
\item Random Case. In this case, we randomly select $k$ labels that are not true labels following a uniform distribution. 
\item Worst Case. In this case, we select $k$ labels that are not true labels with the lowest prediction scores.  These labels are the most difficult to attack. 
\end{compactitem}

\smallskip
\noindent\textbf{Evaluation metrics}. For the instance-specific untargeted and targeted attacks, we use the attack success rate (\asr) as an evaluation metric of the attack performance, which is defined as
\begin{equation} \textstyle
    \mbox{\asr} = 1-\frac{1}{n}\sum_{i=1}^n E(Y(\x_i),\hat{Y}_k(\x_i+\z_i)),
\label{eq:metric_1}
\vspace{-1mm}
\end{equation}
where $n$ is the number of evaluation data. Higher values of \asr~indicate the corresponding method has a high attacking performance. This metric extends the one used in \cite{tursynbek2020geometry} for multi-class classification $|Y|=1$.
In the universal untargted attack, we use a slightly different \asr~definition to reflect that the perturbation is shared by all instances, as $\mbox{\asr} = 1-\frac{1}{n}\sum_{i=1}^n E(Y(\x_i),\hat{Y}_k(\x_i+\z)).$
% \begin{equation} \textstyle
% \label{eq:metric_1}
% \end{equation}  
To evaluate the perceptual quality, we define the average per-pixel perturbation over all successful attacks as
\begin{equation} \textstyle
    \mbox{\pert} = \frac{1}{n \cdot \mbox{\asr}}\sum_{i=1}^n\frac{\|\z_i\|_2(1-E(Y(\x_i),\hat{Y}_k(\x_i+\z_i))}{\mbox{\scriptsize \# of pixels of }\x_i}.
\vspace{-1mm}
\end{equation}
The lower value of \pert~means that the perturbation is less perceivable. The hyper-parameter $\beta$ is chosen to achieve a good trade-off between \asr~and \pert.

\smallskip
\noindent\textbf{Compared Methods}. We use experiments to test the practical performance of T$_k$ML-AP. However, as there are no dedicated adversarial perturbation generation methods for the top-$k$ multi-label learning, we adapt several existing adversarial attacks designed for the general multi-label or multi-class learning as comparison baselines. Specifically, we use the following methods.
\begin{compactitem}
\item Untargeted attack ($k$Fool): We replace the prediction score of one ground-truth label in the $k$Fool \cite{tursynbek2020geometry} algorithm with the maximum prediction score among all ground truth labels as an untargeted attack comparative method. 
\item Universal attack ($k$UAPs): We use the $k$UAPs from \cite{tursynbek2020geometry} with only replace the inner $k$Fool method with our modified untargeted attack comparative method.
\item Targeted attack (ML-AP): We adapt the Rank I algorithm from \cite{song2018multi} with the loss function $[\max_{j \notin P}f_j(\x+\z) - \min_{j\in P}f_{j}(\x+\z)]_+$ to a targeted attack comparative method, where $P$ contains the targeted labels (exclude the ground truth labels) and $|P|=k$. It should be mentioned that this loss is similar to the loss function in \cite{zhang2020learning} when we do not consider the order of targeted labels.
\end{compactitem}
These methods, together with the proposed methods, namely T$_k$ML-AP-U, T$_k$ML-AP-Uv, T$_k$ML-AP-T, are applied to attack the baseline models trained on the datasets. 

%\shu{Instead of taking a long time to find a good trade-off hyper-parameter $\beta$ in all algorithms, we use a projection method \cite{madry2017towards, kurakin2016adversarial, sabour2015adversarial, su2019one} on $\z$. After each iteration, we apply projection on the $\z$ with a projector operation $\mathcal{P}_{\epsilon}$ controls the criteria $\|\z\|_2\leq\epsilon$, where $\epsilon$ is a predefined robustness threshold.} The complete details of the experimental settings can be found in Appendix \ref{append:attack_methods_settings}.
%

\subsection{Results}

\begin{figure}[t]
\captionsetup[subfigure]{font=small}
\centering
\includegraphics[width=\linewidth]{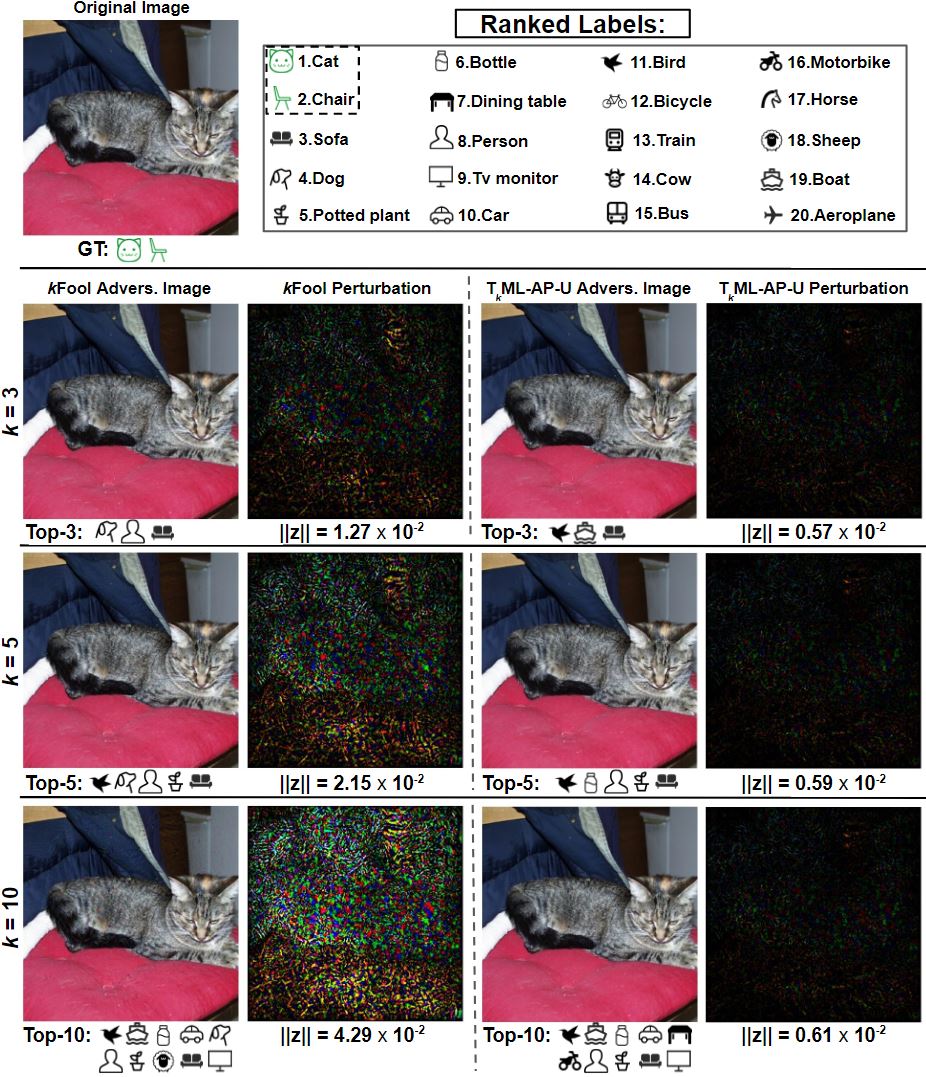}
\vspace{-5mm}
\caption{\small \it Visual examples of untargeted attack methods on PASCAL VOC 2012. The perturbations are scaled by a factor of 20 to increase visibility. Green icons represent the truth labels (GT) that are attacked. The figure is better viewed in color.}\label{fig:UA}
\vspace{-2mm}
\end{figure}

\noindent{\bf Untargeted Attacks}.
\begin{table}[t]
% \captionsetup{font=footnotesize}
\centering
\renewcommand\arraystretch{0.5}
\setlength\tabcolsep{1.5pt}
\scriptsize{
\begin{tabular}{|c|c|cc|cc|}
\hline
\multirow{2}{*}{$k$} & \multirow{2}{*}{Methods} & \multicolumn{2}{c|}{PASCAL VOC 2012} & \multicolumn{2}{c|}{MS COCO 2014} \\ \cline{3-6} 
                  &                   &    \pert($\times$10$^{-2}$)       &    \asr       &     \pert($\times$10$^{-2}$)      &     \asr      \\ \hline
\multirow{2}{*}{3} & $k$Fool& 1.64 &   93.7        &   5.49        &    61.4       \\ 
                  &T$_k$ML-AP-U&   \textbf{0.51}        &   \textbf{99.6}        &    \textbf{0.49}      &    \textbf{100}       \\ \hline
\multirow{2}{*}{5} &$k$Fool&   2.39        &    93.5       &    9.91       &    65.2      \\
                  & T$_k$ML-AP-U&   \textbf{0.56}        &    \textbf{99.3}       &     \textbf{0.53}      & \textbf{100}          \\ \hline
\multirow{2}{*}{10} & $k$Fool &  4.88         &    88.7       &    16.44       &    68.1      \\  
                  & T$_k$ML-AP-U &   \textbf{0.63}        &   \textbf{ 98.3}       &     \textbf{0.59}      &  \textbf{100 }        \\ \hline
\end{tabular}
\vspace{-2mm}
\caption{\small \it Comparison of \pert~and \asr~(\%) of the untargeted attack methods with $k$=3, 5, 10 on two datasets. The best results are shown in bold.}
\label{tab:untargeted-voc}
}
\vspace{-5mm}
\end{table}
The performance of untargeted attacks is shown in Table \ref{tab:untargeted-voc}. Note that for different $k$ values, the T$_k$ML-AP-U method achieves a nearly complete obviation (with very high \asr~values) on both the PASCAL VOC 2012 dataset and the MS COCO 2014 dataset with small perturbation scales (indicated by the smaller \pert~values). On the other hand, the simple adoption of the DeepFool method ($k$Fool) is much less effective. This could be attributed to the explicit consideration of the top-$k$ prediction in T$_k$ML-AP-U. The quantitative results are corroborated with an example from the PASCAL VOC 2012 shown in Fig.\ref{fig:UA}. Although both $k$Fool and T$_k$ML-AP-U show effectiveness in attacking the top-$k$ predictions from the baseline method, $k$Fool introduces larger perturbations in general. In many cases, the perturbations are visible as shown in Fig.\ref{fig:UA}. 

%More experimental and visual results can be found in Appendix \ref{appendix:UA_performance} and \ref{appendix:UA_image_performance}, respectively.

When deployed in practice, it is possible that the attack is designed for top-$k$ predictions but the actual system is used to find the top-$k'$ outputs. In other words, there can be a mismatch between the cutoff rank that is used in the attack $k$ from that used in the system $k'$. Note that by the definition of T$_k$ML-AP-U, a successful attack to a top-$k$ multi-label learning system is necessarily a successful attack to the same system for the case of top-$k'$ for $k' \le k$. This is because the top $k'$ set is a subset of the top $k$ set. On the other hand, we perform a set of experiments to validate the case when $k' > k$. Specifically, in Table \ref{tab:untargeted-other_k}, we show the results of running T$_k$ML-AP-U for $k=3$ and $k^{\prime}=5,10$, respectively.  Note that in these cases, the effectiveness of the effect significantly reduces from the case of $k=k'$. This is expected since a successful top-$k$ attack will move the original top-$k$ labels to ranks greater than $k$. However, our objective Eq.\eqref{e:E3} cannot avoid the case when some of the original labels are placed between $k$ and $k'$, so a successful attack to the top-$k$ case may not generalize to a successful attack to the case of top-$k', (k < k')$. 

\begin{table}[t]
% \captionsetup{font=footnotesize}
\centering
\setlength\tabcolsep{1.5pt}
\scriptsize{
\begin{tabular}{|c|c|cc|cc|}
\hline
\multirow{2}{*}{$k^{\prime}$} & \multirow{2}{*}{\makecell{Method\\ ($k$=3)}} & \multicolumn{2}{c|}{PASCAL VOC 2012} & \multicolumn{2}{c|}{MS COCO 2014} \\ \cline{3-6} 
                  &                   &    \pert($\times$10$^{-2}$)       &    \asr       &     \pert($\times$10$^{-2}$)      &     \asr      \\ \hline
3 & T$_k$ML-AP-U&   0.51        &   99.6        &    0.49      &    100        \\ \hline
5 & T$_k$ML-AP-U&   0.24        &   3.6        &    0.43       &    26.5       \\ \hline
10 & T$_k$ML-AP-U&   0.18        &    0.3      &     0.35      & 3.9          \\ \hline
% \multirow{2}{*}{5} & $k$Fool& 3.72 &   15.7        &   3.89        &    5.5       \\ 
%                   &T$_k$ML-AP-U&   3.36        &   45.3        &    24.61       &    82.1       \\ \hline
% \multirow{2}{*}{10} &$k$Fool&   3.57        &   3.2       &    2.53       &    0.4      \\
%                   & T$_k$ML-AP-U&   3.38        &    15.4       &     24.89      & 55.4          \\ \hline
\end{tabular}
\vspace{-1mm}
\caption{\small \it Comparison of \pert~and \asr~(\%) of the untargeted attack methods in $k^{\prime}=3, 5,10$ on two datasets when setting $k$=3.}
\label{tab:untargeted-other_k}
}
\vspace{-3mm}
\end{table}

\begin{table}[t]
% \captionsetup{font=footnotesize}
\centering
\setlength\tabcolsep{1.5pt}
\scriptsize{
\begin{tabular}{|c|cc|cc|cc|}
\hline
         $k$ & \multicolumn{2}{c|}{1} & \multicolumn{2}{c|}{2} & \multicolumn{2}{c|}{3} \\ \hline
            Metrics&    \pert &  \asr & \pert &  \asr & \pert &  \asr         \\ \hline
           $k$UAPs &   0.51  & 63.9&0.51 & 74.6 &0.51 &73.2           \\ 
           T$_k$ML-AP-Uv&   \textbf{0.13} & \textbf{86.5} &\textbf{0.15}  &\textbf{82}&\textbf{0.16}&\textbf{ 80.5 }  \\ \hline
\end{tabular}
\vspace{-1mm}
}
\caption{\small \it Comparison of \pert~and \asr~(\%) of the universal untargeted attack methods on MS COCO 2014.}
\label{tab:uap_ep_100_xi_70}
\vspace{-5mm}
\end{table}

\begin{figure}[t]
\captionsetup[subfigure]{font=small}
\centering
\includegraphics[width=\linewidth]{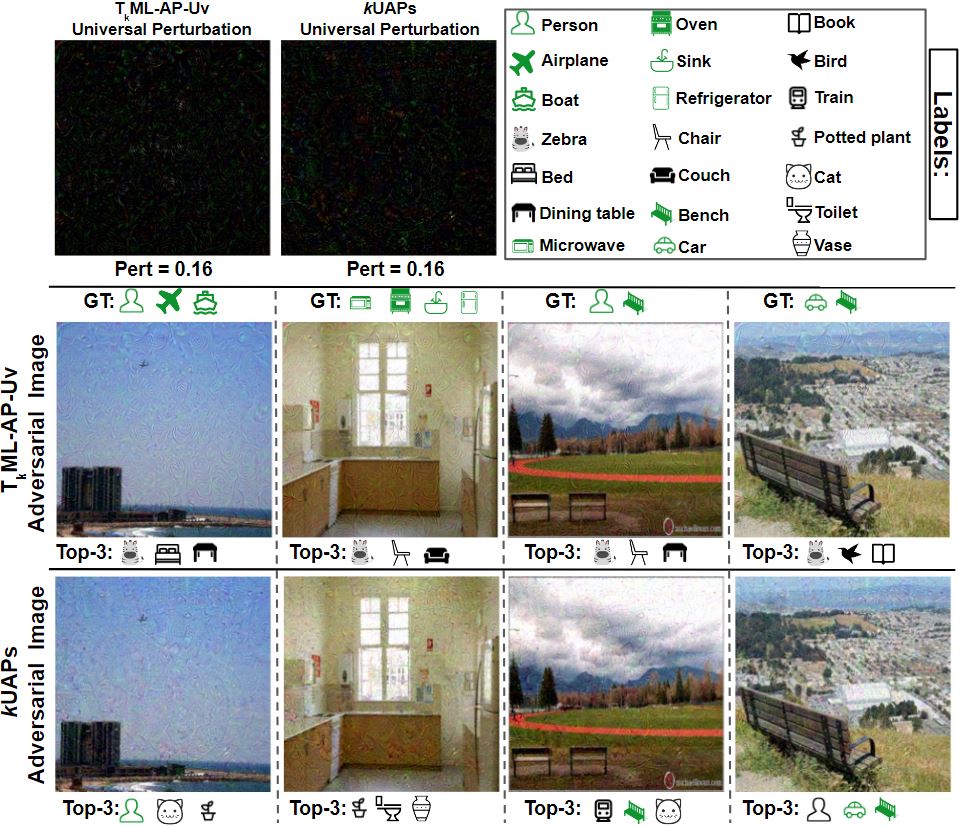}
\vspace{-5mm}
\caption{\small \it Examples of universal untargeted attack methods with $k=3$ on MS COCO 2014. The figure is better viewed in color.}\label{fig:UUA}
\vspace{-2mm}
\end{figure}

\smallskip
\noindent{\bf Universal Untargeted Attacks}. 
The results of universal untargeted attacks are shown in Table \ref{tab:uap_ep_100_xi_70}.  On the MS COCO 2014 dataset, T$_k$ML-AP-Uv outperforms $k$UAPs in all cases. Fig.\ref{fig:UUA} further exhibits visual examples of universal perturbation with T$_k$ML-AP-Uv and $k$UAPs for $k=3$. With similar perturbations,  T$_k$ML-AP-Uv is successful in attacking all top-$3$ labels but there are images that $k$UAPs fails to attack. On the other hand, because of the requirement of being instance-independent, to achieve the same level of attacks, universal untargeted attacks need to introduce larger visual perturbations than those in the instance-specific attacks. 

%Additional quantitative performances and visual results can be found in Appendix \ref{appendix:UAP_performance} and \ref{appendix:UAP_image_performance}.

\begin{table}[t]
% \captionsetup{font=footnotesize}
\centering
\renewcommand\arraystretch{0.5}
\setlength\tabcolsep{3.5pt}
\scriptsize{
\begin{tabular}{|c|c|c|cc|cc|}
\hline
\multirow{2}{*}{Cases} & \multirow{2}{*}{$k$} & \multirow{2}{*}{Methods} & \multicolumn{2}{c|}{PASCAL VOC 2012} & \multicolumn{2}{c|}{MS COCO 2014} \\ \cline{4-7} 
                  &                   &                  &\pert($\times$10$^{-2}$)&\asr & \pert($\times$10$^{-2}$) &\asr           \\ \hline
\multirow{6}{*}{Best} & \multirow{2}{*}{3} & ML-AP& 0.44 & 96.2&0.55 &100   \\ 
                  &                   & T$_k$ML-AP-T &0.44&\textbf{96.6}&0.57&\textbf{100} \\ \cline{2-7} 
                  & \multirow{2}{*}{5} & ML-AP &0.50&92 &0.66&99.9\\
                  &                   &T$_k$ML-AP-T&0.50 & \textbf{92.8 }&0.69 & \textbf{99.9}\\ \cline{2-7} 
                  & \multirow{2}{*}{10} & ML-AP &0.55& 84.2&0.81 &99.8 \\  
                  &                   & T$_k$ML-AP-T &0.56&\textbf{86.4}&0.85 & \textbf{99.8}\\ \hline
\multirow{6}{*}{Random} & \multirow{2}{*}{3} & ML-AP &0.59 &86 &0.95 &99.8 \\ 
                  &                   & T$_k$ML-AP-T &0.59&\textbf{89.8} &0.99 &\textbf{99.9} \\ \cline{2-7} 
                  & \multirow{2}{*}{5} & ML-AP&0.62&77.9& 1.11&96.5 \\ 
                  &                   &T$_k$ML-AP-T&0.63&\textbf{83.7}&1.18  &\textbf{97.8} \\ \cline{2-7} 
                  & \multirow{2}{*}{10} &ML-AP &0.63&67.7&1.22& 84.2 \\ 
                  &                   &T$_k$ML-AP-T & 0.64&\textbf{76.4}&1.28  &\textbf{94.5}\\ \hline
\multirow{6}{*}{Worst} & \multirow{2}{*}{3} &ML-AP & 0.66& 68&1.08 & 90\\ 
                  &                   & T$_k$ML-AP-T &0.66&\textbf{75.8} &1.14 &\textbf{91.4}\\ \cline{2-7} 
                  & \multirow{2}{*}{5} & ML-AP&0.67 &53.3&1.18 &81.8 \\  
                  &                   & T$_k$ML-AP-T&0.69&\textbf{66.6}&1.25&\textbf{87.2} \\ \cline{2-7} 
                  & \multirow{2}{*}{10} &ML-AP & 0.67&39.1&1.25 & 59 \\
                  &                   & T$_k$ML-AP-T & 0.69&\textbf{57} &1.30& \textbf{73.1} \\ \hline
\end{tabular}
\vspace{-1mm}
}
\caption{\small \it Comparison of \pert~and \asr~(\%) of the targeted attack methods with $k$=3, 5, 10 in the Best, Random, and Worst cases on two datasets. The best \asr~results are shown in bold.}
\label{tab:targeted_ep_2}
\vspace{-5mm}
\end{table}

\begin{figure}[t]
\captionsetup[subfigure]{justification=centering}
\centering
\includegraphics[width=1\linewidth]{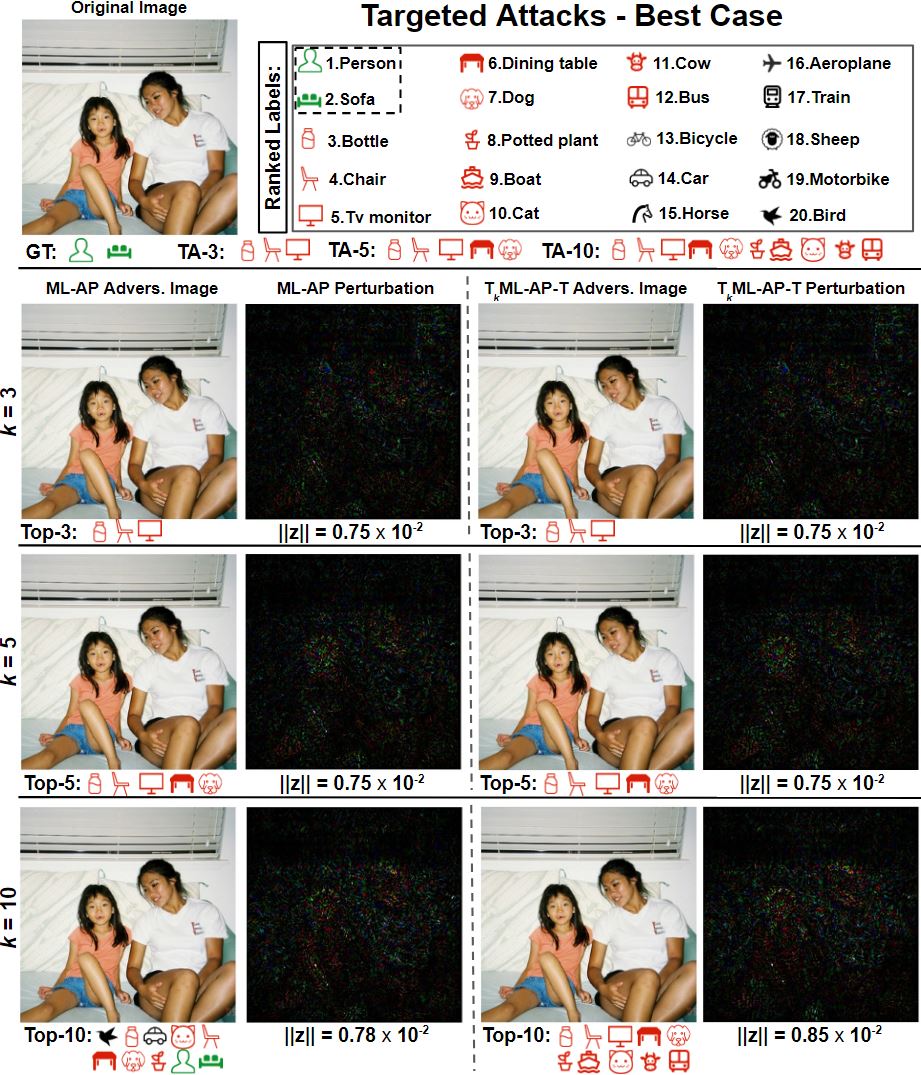}\\
% \vspace{0.3mm}
% \smallskip
\includegraphics[width=1\linewidth]{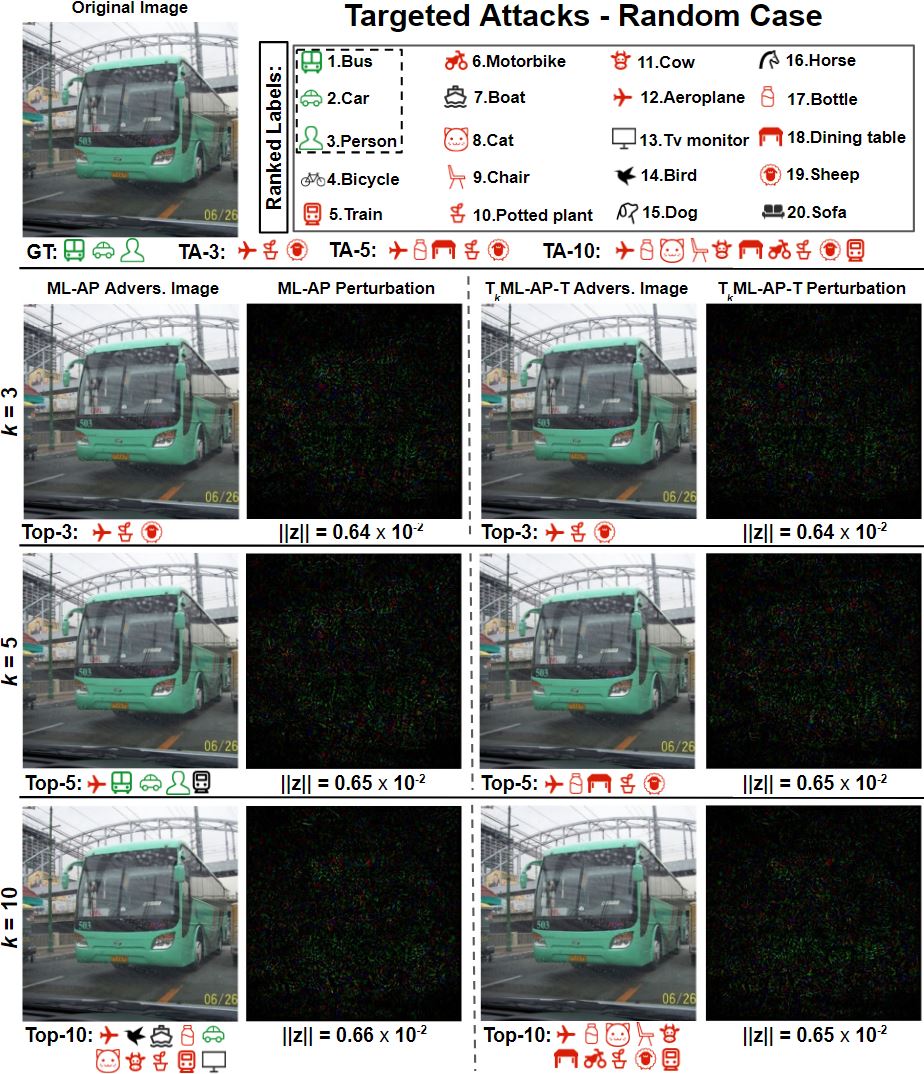}
\vspace{-5mm}
\caption{\small \it Targeted attack in Best (\textbf{left-top}, targeted labels are near GT) , Random (\textbf{left-bottom}, targeted labels are randomly selected), and Worst (\textbf{right}, targeted labels are far from GT) cases. TA means the targeted attack labels. The perturbations are scaled by a factor of 20 to increase visibility. The red icons represent the targeted labels for attacking. The figure is better viewed in color.}
\label{fig:TA}
\vspace{-10mm}
\end{figure}

\smallskip
\noindent{\bf Targeted Attacks}.
In Table \ref{tab:targeted_ep_2}, we evaluate the performance of  T$_k$ML-AP-T (our method) and compare it with the ML-AP method in the three attack settings (Best, Random, and Worst) as described in Section \ref{sec:exp-set}.  On both datasets and over all cases, T$_k$ML-AP-T outperforms ML-AP for different $k$ values and comparing settings in terms of the \asr~scores with comparable perturbation strengths. In particular, with increasing $k$, the gap of \asr~scores between T$_k$ML-AP-T and the ML-AP method also increases.  On the other hand, we note that the \asr~scores decrease when the $k$ value increases. This is because the attack methods need to take more effort to put labels in the set $\widetilde{Y}$ to top-$k$ positions. The attacks become more challenging for both methods when the target labels are chosen according to the Worst setting, reflecting that larger perturbations are required to modify the predictions to the more difficult labels. Visual results of targeted attack using T$_k$ML-AP-T and ML-AP are shown in Fig.\ref{fig:TA}. 

% \begin{figure}[t]
% \captionsetup[subfigure]{justification=centering}
% \centering
% % \includegraphics[width=\linewidth]{LaTeX/figs/TA_ith_277.jpg}
% \includegraphics[width=\linewidth]{LaTeX/figs/TA_ith_247.jpg}
% % \includegraphics[width=\linewidth]{LaTeX/figs/TA_ith_647.JPG}
% \vspace{-1mm}
% \caption{\small \it see Caption of Fig xxx for more information.}\label{fig:TA_2}
% \vspace{-5mm}
% \end{figure}

\begin{figure}[t]
\captionsetup[subfigure]{justification=centering}
\centering
\includegraphics[width=\linewidth]{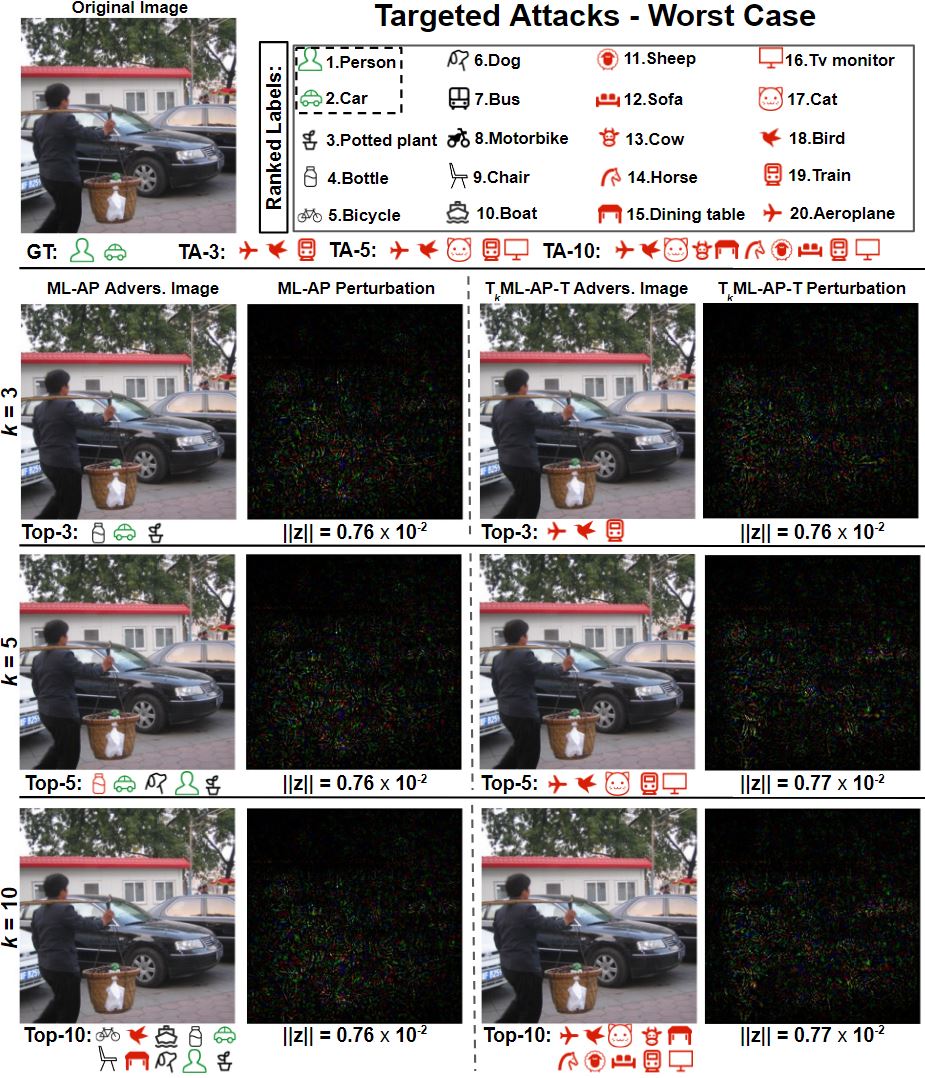}
\vspace{-4mm}
% \caption{\small \it See Caption of Fig.\ref{fig:TA_3} for more information.}\label{fig:TA_3}
\vspace{-7mm}
\end{figure}

%The image results on the Best case and the Worst case can be found in Appendix \ref{appendix:TA_image_performance}.  From Fig.\ref{fig:TA}, we can see that both targeted attack methods have the same performance when $k=3$. However, when $k$ increases, the  top-$k$ predicted results from the ML-AP method contain  truth labels. In contrast, our method can put all predefined targeted attack labels in the top-$k$ position, while keeping the perturbation norm similar to the norms from the ML-AP method. 

\section{Conclusion}
%\xin{is this section too long?}

Top-$k$ multi-label learning (T$_k$ML) has many practical applications. However, the vulnerability of such algorithms with regards to dedicated adversarial perturbation attacks has not been extensively studied previously. In this work, we develop white-box generation methods of adversarial perturbations to T$_k$ML based image annotation system for both untargeted and targeted attacks. Our methods explicitly consider the top-$k$ ranking relation and are based on novel loss functions. Experimental evaluations on large-scale benchmark datasets including PASCAL VOC and MS COCO demonstrate the effectiveness of our methods in reducing the performance of state-of-the-art T$_k$ML methods. 

There are several directions in which we would like to further improve our current methods. First, we only consider white-box attacks in the current work, it is natural to extend similar attacks to the black-box setting in which we do not have detailed knowledge of the model to be attacked. Furthermore, labels are not independent, and targeted attacks tend to be easier for semantically distant labels, \eg, it is probably easier to change label {\tt Dog} to {\tt Cat} than {\tt Airplane}. Therefore, in our subsequent work, we would like to consider the semantic dependencies in designing more effective attacks to T$_k$ML algorithms. 
% More importantly, our results expose the potential vulnerabilities of existing multi-label learning methods and can be used to improve them. 
We will also study defenses against such attacks as an important future work.

\smallskip
\noindent {\bf Acknowledgments}. This research was developed with funding from the National Science Foundation under Grant No. IIS-2103450.

% must be 8 pages
%\newpage

{\small
\bibliographystyle{ieee_fullname}
\bibliography{main}
}

\newpage

\input{Appendix}

\end{document}

%% file: Appendix.tex
\appendix
\def\p{\mathbf{p}}
\def\v{\mathbf{v}}
\def\u{\mathbf{u}}

{\bf \Large  Appendix}

\bigskip 

\section{Proofs}
\subsection{Proof of Lemma 1} \label{append:proof_lemma_1}
\begin{proof}
We know $\sum_{j=1}^k f_{[j]}$ is the solution of 
\begin{equation*}
    \max_\p \p^\top F, ~\text{s.t.}~ \p^\top \mathbf{1}=k, \mathbf{0}\leq\p\leq \mathbf{1}.
\end{equation*}
We apply Lagrangian to this equation and get
\begin{equation*}
    L = -\p^\top F-\v^\top\p+\u^\top(\p-1)+\lambda(\p^\top\mathbf{1}-k)
\end{equation*}
where $\u\geq\mathbf{0}$, $\v\geq\mathbf{0}$ and $\lambda\in\mathbb{R}$ are Lagrangian multipliers. Taking its derivative w.r.t. $\p$ and set it to 0, we have $\v=\u-F+\lambda\mathbf{1}$. Substituting it back into the Lagrangian, we get 
\begin{equation*}
    \min_{\u,\lambda} \u^\top \mathbf{1}+k\lambda, ~\text{s.t.}~ \u\geq\mathbf{0}, \u+\lambda\mathbf{1}-F\geq 0.
\end{equation*}
This means 
\begin{equation*}
    \sum_{j=1}^k f_{[j]}=\min_{\lambda}\Big\{k\lambda+\sum_{j=1}^m[f_j-\lambda]_+\Big\}.
\end{equation*}
Therefore, 
\begin{equation}
    \phi_k(F)=\frac{1}{k}\min_{\lambda}\Big\{k\lambda+\sum_{j=1}^m[f_j-\lambda]_+\Big\}.
\label{eq:topk_convex}
\end{equation}
Furthermore, we can see that $\lambda=f_{[k]}$ is always one optimal solution for Eq.(\ref{eq:topk_convex}). So
\begin{equation*}
    f_{[k]}\in\arg\min_{\lambda}\Big\{k\lambda+\sum_{j=1}^m[f_j-\lambda]_+\Big\}.
\end{equation*}
\end{proof}

\subsection{Proof of Lemma 2} \label{append:proof_lemma_2}
\begin{proof}
Denote $g(x)=[[a-x]_+-\lambda]_+$. For $\lambda\geq 0$, we have $g(x)=0=[a-x-\lambda]_+$ if $x\geq a$. On the other hand, if $x<a$, we have $g(x)=[a-x-\lambda]_+$. Therefore, $g(x)=[[a-x]_+-\lambda]_+= [a-x-\lambda]_+$ for any $\lambda\geq 0$.
\end{proof}

\section{Additional Experimental Details} \label{append:add_exp_details}

\subsection{Source Code}
For the purpose of review, the source code is accessible in the supplementary file.

\subsection{Computing Infrastructure Description}
All algorithms are implemented in Python 3.6 and trained and tested on an Intel(R) Xeon(R) CPU W5590 @3.33GHz with 48GB of RAM and an NVIDIA Quadro RTX 6000 GPU with 24GB memory.

\subsection{The Algorithm for The Universal Untargeted Attack} \label{append:UUA_algorithm}
First, we define the universal attack success rate (\uasr) as
\begin{equation}
    \begin{aligned}
    \mbox{\uasr}&=\frac{1}{n}\sum_{i=1}^n\mathbb{I}_{\left[E(Y(\x_i),\hat{Y}(\x_i+\z)) = 0\right]}.
    \end{aligned}
\label{eq:uap_metric}
\end{equation}
We apply Algorithm \ref{Alg1} (without using projection in it) iteratively over all samples from $\textbf{X}$. At each iteration, Algorithm \ref{Alg1} finds a perturbation $\Delta \z_i$ for a given data point $\x_i+\z$, which success attack all top-$k$ labels for the current data $x_i$. Then we update the universal adversarial perturbation $\z$ by using a projection operation to it. The overall procedure is described in detail in Algorithm \ref{Alg2}. The algorithm terminates when the predefined attack success rate $\xi$ is reached.  
\begin{algorithm}[ht]
    \caption{T$_k$ML-AP-Uv}\label{Alg2}
    \SetAlgoLined
    % \begin{flushleft}
    \KwIn{$\textbf{X}=\{\x_1,\cdots,\x_n\}$, predictor $F$, $k$, $\eta_l$, $\xi$, $\beta$}
    \KwOut{perturbation $\z^*$} 
    % \textbf{Input:} $\textbf{x}$, predictor $F$, set $P$, maxiter   \\
    % \textbf{Output:} adversarial example $\textbf{x}^*$, perturbation $\z^*$  \\
    \textbf{Initialization:} $\z$ \\

    % \end{flushleft}
    
    \While{$\mbox{\uasr} < \xi$}{
    \For{$\x_i \in \textbf{X}$}{
    \If{$E(Y(\x_i),\hat{Y}(\x_i+\z)) \neq 0$}{
    % \While {$l\leq $ maxiter}{
    $\_ \ ,\Delta\z_{i} = $T$_k$ML-AP-U($\x_i+\z,F,k,\eta_l,\beta$) $\triangleright$ Algorithm \ref{Alg1}

    % }
    $\z = \mathcal{P}_{\epsilon}(\z+\Delta\z_{i})$
    }
    }
    }
    $\z^* = \z$
    
    \Return{$\z^*$}
\end{algorithm}

\subsection{Baseline Model Settings}\label{append:baseline_methods_settings}
Specifically, we tune the model with Adam optimizer with an initial learning rate of $0.001$ and batch size of 64 on PASCAL VOC 2012 and MS COCO 2014 for 25 and 100 epochs respectively. 

For the PASCAL VOC 2012 dataset, following the protocol of \cite{song2018multi, zhou2020generating} for a fair comparison, which trained on the training set (5,717 images) and tests on the validation set (5,823 images). The MS COCO 2014 dataset is a larger dataset comparing with the PASCAL VOC 2012 in terms of both numbers of classes and images. It does not provide the ground truth labels for the testing images either. Similarly, we do the training on the training set (82,081 images) and testing on the validation set (40,137 images). The images are normalized into the range of $[-1, 1]$.

\begin{table*}[th!]
% \captionsetup{font=footnotesize}
\centering
\setlength\tabcolsep{1.5pt}
\scriptsize{
\begin{tabular}{|c|c|cc|cc|cc|cc|cc|cc|}
\hline
\multirow{3}{*}{$k$} & \multirow{3}{*}{Methods} & \multicolumn{6}{c|}{PASCAL VOC 2012}                                                 & \multicolumn{6}{c|}{MS COCO 2014}                                                 \\ \cline{3-14} 
                  &                   & \multicolumn{2}{c|}{$k^{\prime}$=3} & \multicolumn{2}{c|}{$k^{\prime}$=5} & \multicolumn{2}{c|}{$k^{\prime}$=10} & \multicolumn{2}{c|}{$k^{\prime}$=3} & \multicolumn{2}{c|}{$k^{\prime}$=5} & \multicolumn{2}{c|}{$k^{\prime}$=10} \\ \cline{3-14} 
                  & & \pert($\times$10$^{-2}$)  & \asr & \pert($\times$10$^{-2}$)  & \asr & \pert($\times$10$^{-2}$) &\asr&\pert($\times$10$^{-2}$) &\asr &\pert($\times$10$^{-2}$)&\asr& \pert($\times$10$^{-2}$) &\asr           \\ \hline
\multirow{2}{*}{3} & $k$Fool &1.64& 93.7 &0.95 &15.7&0.91 & 3.2&5.48&61.4&0.99&5.5 &0.65&0.4 \\ 
                  &T$_k$ML-AP-U&\textbf{0.51}&\textbf{99.6}& 0.24 & 3.6&0.18 &0.3& \textbf{0.24} &\textbf{100}&0.43&26.5& 0.35&3.9\\ \hline
\multirow{2}{*}{5} & $k$Fool &- &-&2.39 &93.5&1.59&20.4&- &- & 9.92 &65.2& 1.10&6.8 \\ 
                  &T$_k$ML-AP-U &-&-&\textbf{0.56}&\textbf{99.3}&0.18 &1&- &- &\textbf{0.53} & \textbf{100} & 0.39&9.8\\ \hline
\multirow{2}{*}{10} &$k$Fool &-&-&-&-&4.87&88.7&- &-&-& -& 1.66 & 68.1 \\
                  & T$_k$ML-AP-U &-&- & -&-& \textbf{0.63} &\textbf{98.3}&-&-&-&- &\textbf{0.59} &\textbf{100}  \\ \hline
\end{tabular}
\vspace{-3mm}
\caption{\small \it Comparison of \pert~and \asr~(\%) of the untargeted attack methods with $k=3,5,10$ on two datasets. The best results are shown in bold. `-' represents the current results in the $k^{\prime}<k$ setting are the same as the results in the $k^{\prime}=k$ setting.}
\label{tab:untargeted-eps_10}
}
\vspace{-5mm}
\end{table*}
\subsection{Settings of Attacking Methods}\label{append:attack_methods_settings}
We use the same learning rate $0.01$ for T$_k$ML-AP-U, T$_k$ML-AP-T, and ML-AP methods. We set 1000 as the maximum iterations in all untargeted and targeted attack methods.
we only test the $\ell_2$ norm for the perturbations. However, our algorithms can also work on other $\ell_p$ norms, \ie, $\ell_1$ and $\ell_\infty$. 
Since the optimization may get stuck in extreme spots, we follow similar image processing and variable transformation methods in the algorithms based on \cite{carlini2017towards} to avoid this problem. 
% These methods may cause the final perturbation slightly large than the predefined threshold when using projection. However, it does not influence our performance comparison in the experiments. 

Instead of taking a long time to find a good trade-off hyper-parameter $\beta$ in all algorithms, we use a projection method 
% \cite{madry2017towards, kurakin2016adversarial, sabour2015adversarial, su2019one} 
\cite{madry2017towards, kurakin2016adversarial} 
on $\z$. After each iteration, we apply projection on the $\z$ with a projector operation $\mathcal{P}_{\epsilon}$ controls the criteria $\|\z\|_2\leq\epsilon$, where $\epsilon$ is a predefined robustness threshold.

In the untargeted attack experiments, we set the projection threshold $\epsilon=10$ in our algorithm. Since our algorithm and the $k$Fool method have no terminate conditions, they will success attack all images in the test set of data. This means both of them can achieve a 100\% \asr~score in the final performance even take a long time in some specific images. To avoid this situation, we set the maximum iteration equals 1000 as we mentioned before. After both algorithms finish attack 1000 images in each dataset, we report the final performance.

We set $\epsilon=2$ in the targeted attack experiment and report the final performance in the main paper. However, we also test other $\epsilon$ settings in the following Section.

\section{Additional Experimental Results}
\subsection{Additional Untargeted Attacking Performance}\label{appendix:UA_performance}

\begin{table}[t]
% \captionsetup{font=footnotesize}
\centering
\scriptsize{
\begin{tabular}{|c|c|c|c|}
\hline
\multirow{2}{*}{Methods} & \multirow{2}{*}{$k$} & \multicolumn{2}{c|}{MS COCO 2014} \\ \cline{3-4} 
                  &                   &      \pert($\times$10$^{-2}$)     &    \asr       \\ \hline
\multirow{6}{*}{$k$Fool} &10&16.45&  68.1         \\ \cline{2-4} 
                  &15 & 23.49 & 66.7     \\ \cline{2-4} 
                  &20 & 26.01 &65.7      \\ \cline{2-4} 
                  &25& 57.38 & 63    \\ \cline{2-4} 
                  &30& 97.67 &62.7 \\ \cline{2-4} 
                  &40& 103.53 & 51.2 \\ \hline
\end{tabular}
\vspace{-3mm}
\caption{\small \em \pert~and \asr~(\%) of $k$Fool method in different $k$ settings on MS COCO 2014 dataset.}
\label{tab:untargeted_baseline_large_K}
}
\vspace{-5mm}
\end{table}

\begin{figure}[t]
\captionsetup[subfigure]{justification=centering, font=small}
\centering
\includegraphics[width=\linewidth]{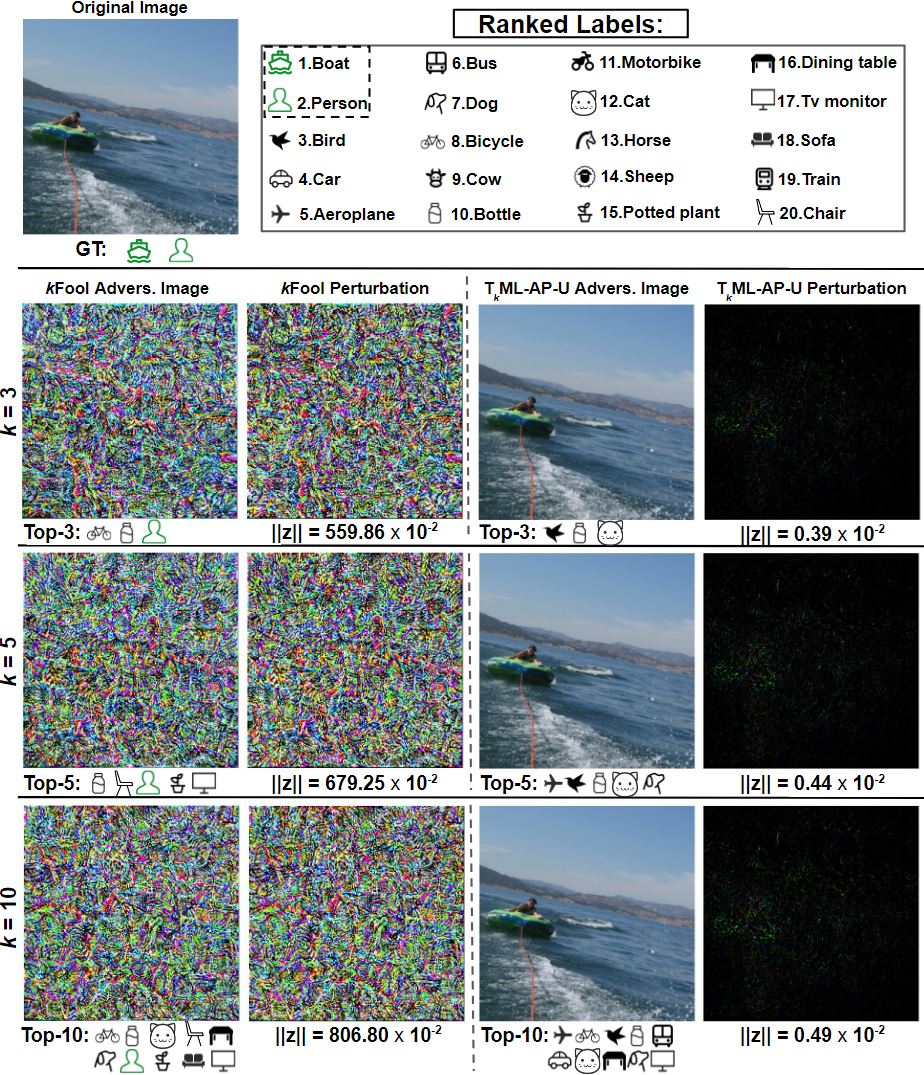}
\vspace{-5mm}
\caption{\small \it Examples of $k$Fool and T$_k$ML-AP-U adversarial perturbations with $k=3, 5, 10$ on PASCAL VOC 2012. To better show perturbations, we have multiplied the intensity of all perturbation images by 20. GT means the ground truth labels. Top-$k$ ($k$=3, 5, 10) means the Top-$k$ predicted labels from the corresponding image. Advers. means adversarial. The green icons correspond to the ground truth labels.}\label{fig:UA_fail_case}
\vspace{-5mm}
\end{figure}

First, we report the complete results with the same setting from Table \ref{tab:untargeted-voc} in Table \ref{tab:untargeted-eps_10}. Second, we analysis some results displayed in Table \ref{tab:untargeted-voc}. For the $k$Fool method, we can see that \asr~scores from 93.7\% to 88.7\%, which are decreasing with increasing the $k$ value in the PASCAL VOC 2012 dataset. This is because the top-$k$ attack becomes hard when the $k$ is increasing. The algorithm needs to take more effort to push ground truth labels outside the top-$k$ position. However, we find an opposite trend that \asr~score is increasing when the $k$ value is increasing in the COCO dataset. The reason is that the number of labels in the COCO dataset is four times that in VOC. In other words, the number of non-ground truth labels is very large in the COCO dataset. When the $k$ value is small, the performance of the $k$Fool method is not influenced much by the $k$ settings. However, when the $k$ value is large and continues increasing, the \asr~score will be decreased because the $k$ value has more impact on the algorithm. We have verified this statement through additional experiments in Table \ref{tab:untargeted_baseline_large_K}.

\subsection{Additional Untargeted Attacking Image Results}\label{appendix:UA_image_performance}

\begin{figure}[t]
\captionsetup[subfigure]{justification=centering, font=small}
\centering
\includegraphics[width=\linewidth]{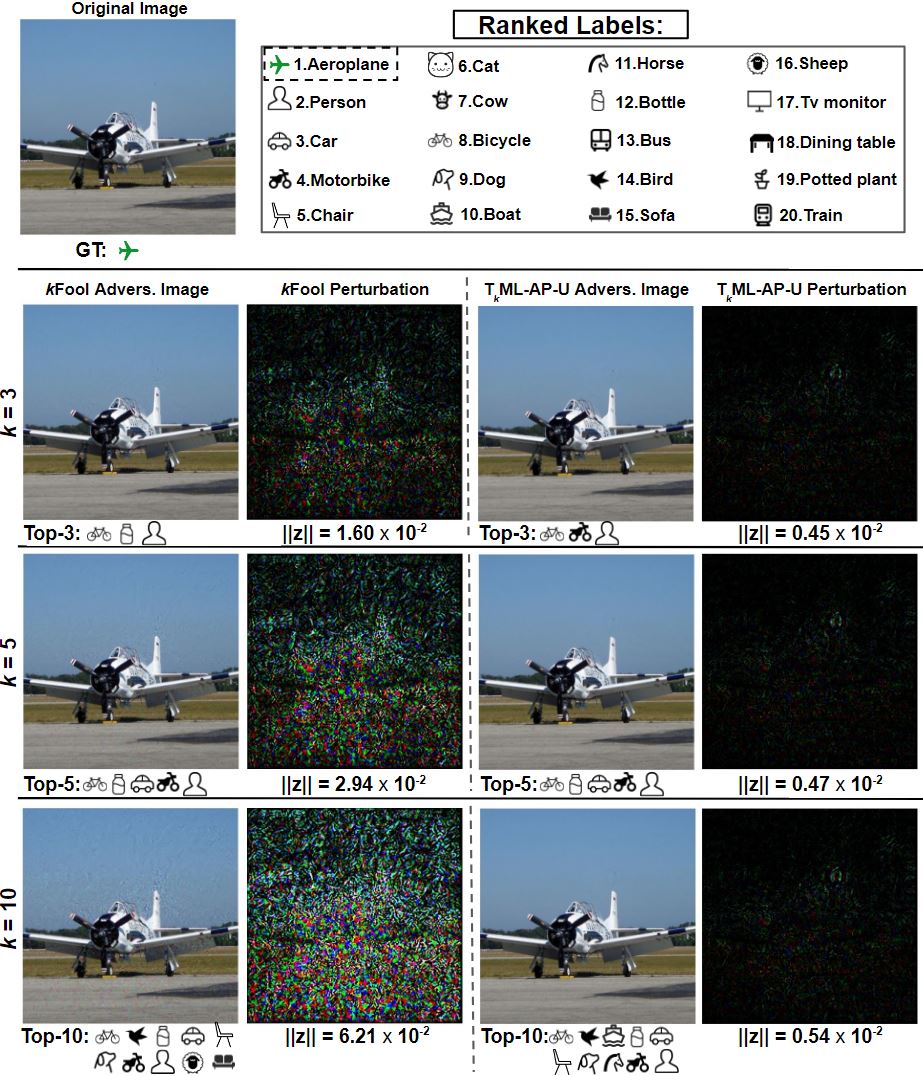}
\vspace{-5mm}
\caption{\small \it Examples of $k$Fool and T$_k$ML-AP-U adversarial perturbations with $k=3, 5, 10$ on PASCAL VOC 2012. To better show perturbations, we have multiplied the intensity of all perturbation images by 20. GT means the ground truth labels. Top-$k$ ($k$=3, 5, 10) means the Top-$k$ predicted labels from the corresponding image. Advers. means adversarial. The green icons correspond to the ground truth labels.}\label{fig:UA_case_5}
\vspace{-5mm}
\end{figure}

We show a failed example from  the $k$Fool method in Figure \ref{fig:UA_fail_case}. Since the original $k$Fool method is for attacking the top-$k$ multi-class classifier, we show the results, which an image has only one ground-truth label in Figure \ref{fig:UA_case_5}. From Figure \ref{fig:UA_case_5}, we can see that our method outperforms the $k$Fool method even our method is reduced to a multi-class version. 

\subsection{Additional Universal Untargeted Attacking Performance} \label{appendix:UAP_performance}

We also compare the performance of our T$_k$ML-AP-Uv method and the $k$UAPs method. In the training of both algorithms, when \uasr~larger than 0.7, we output the universal perturbation $\|\z\|$ and use it to evaluate the attacking methods and report the performance in Table \ref{tab:uap_ep_100_xi_70_old}. Note that the performance in Table \ref{tab:uap_ep_100_xi_70} is a partial results of the performance in Table \ref{tab:uap_ep_100_xi_70_old}. To evaluate the performance efficiently and avoid the algorithm still get stuck in the loop, we set 20 as the maximum iteration for the outer loop in both algorithms. We use `$\times$' to represent the algorithm that cannot satisfy the terminate conditions after the maximum iteration. We also report the results when $\xi =0.8$.

\begin{table}[t]
% \captionsetup{font=footnotesize}
\centering
\renewcommand\arraystretch{0.5}
\setlength\tabcolsep{1.5pt}
\scriptsize{
\begin{tabular}{|c|c|c|c|c|c|c|c|c|c|c|}
\hline
\multirow{2}{*}{$\xi$} &\multirow{2}{*}{$k$} & \multirow{2}{*}{Methods} & \multicolumn{4}{c|}{PASCAL VOC 2012} & \multicolumn{4}{c|}{MS COCO 2014} \\ \cline{4-11} 
        &          &                   &   $k^{\prime}$=1  &  $k^{\prime}$=2   &  $k^{\prime}$=3   & \pert   &   $k^{\prime}$=1  &  $k^{\prime}$=2   &  $k^{\prime}$=3   & \pert   \\ \hline
\multirow{6}{*}{0.7}&\multirow{2}{*}{1} & $k$UAPs&  $\times$&	$\times$&	$\times$&	$\times$&	63.9&	51.4&	45&	0.51   \\ 
                  &&T$_k$ML-AP-Uv&\textbf{ 72.3}&	60&	50.2&\textbf{0.15}&	\textbf{86.5}&	68.5&	62.9&	\textbf{0.13} \\ \cline{2-11} 
&\multirow{2}{*}{2} & $k$UAPs & -&	$\times$&	$\times$&	$\times$&	-&	74.6&	65.9&	0.51 \\ 
                 & &  T$_k$ML-AP-Uv  & -&	\textbf{68}&	59.5&	\textbf{0.16}&	-	&\textbf{82}&	82&	\textbf{0.15}  \\ \cline{2-11} 
&\multirow{2}{*}{3} & $k$UAPs &-	&-&	$\times$&	$\times$&	-&	-&	73.2&	0.51  \\ 
                 & &T$_k$ML-AP-Uv& -&	-&	\textbf{65}&	\textbf{0.17}&	-&	-&\textbf{	80.5}&	\textbf{0.16}\\ \hline
\multirow{6}{*}{0.8}&\multirow{2}{*}{1} & $k$UAPs&  $\times$&	$\times$&	$\times$&	$\times$&	66.2&	56.2&	49.4&	0.51   \\ 
                  &&T$_k$ML-AP-Uv&\textbf{ 74.2}&	60.7&	50.4&\textbf{0.14}&	\textbf{84.5}&	70.4&	63.7&	\textbf{0.13} \\ \cline{2-11} 
&\multirow{2}{*}{2} & $k$UAPs & -&	$\times$&	$\times$&	$\times$&	-&	$\times$&	$\times$&	$\times$ \\ 
                 & &  T$_k$ML-AP-Uv  & -&	\textbf{70.5}&	61.7&	\textbf{0.16}&	-	&\textbf{81.2}& 75.3&	\textbf{0.15}  \\ \cline{2-11} 
&\multirow{2}{*}{3} & $k$UAPs &-	&-&	$\times$&	$\times$&	-&	-&	73.2&	0.51  \\ 
                 & &T$_k$ML-AP-Uv& -&	-&	\textbf{69}&	\textbf{0.18}&	-&	-&\textbf{	78.9}&	\textbf{0.16}\\ \hline
\end{tabular}
\vspace{-1mm}
}
\caption{\small \it Comparison of \pert~and \asr~(\%) of the universal untargeted attack methods for two datasets. `$\times$' represents the current algorithm cannot get results. `-' represents the current results in the $k^{\prime}<k$ setting are the same as the results in the $k^{\prime}=k$ setting. $\epsilon=100$. The best results are shown in bold.}
\label{tab:uap_ep_100_xi_70_old}
\vspace{-3mm}
\end{table}

\begin{table}[t]
% \captionsetup{font=footnotesize}
\centering
\renewcommand\arraystretch{0.5}
\setlength\tabcolsep{1.5pt}
\scriptsize{
\begin{tabular}{|c|c|c|c|c|c|c|c|c|c|c|}
\hline
\multirow{2}{*}{$\xi$} &\multirow{2}{*}{$k$} & \multirow{2}{*}{Methods} & \multicolumn{4}{c|}{PASCAL VOC 2012} & \multicolumn{4}{c|}{MS COCO 2014} \\ \cline{4-11} 
        &          &                   &   $k^{\prime}$=1  &  $k^{\prime}$=2   &  $k^{\prime}$=3   & \pert    &   $k^{\prime}$=1  &  $k^{\prime}$=2   &  $k^{\prime}$=3   & \pert   \\ \hline
\multirow{6}{*}{0.7}&\multirow{2}{*}{1} & $k$UAPs&  $\times$&	$\times$&	$\times$&	$\times$&	\textbf{97.6}&	91.5&	85.4&	1.24   \\ 
                  &&T$_k$ML-AP-Uv&\textbf{ 72.3}&	60&	50.2&\textbf{0.14}&	86.5&	68.5&	62.9&	\textbf{0.13} \\ \cline{2-11} 
&\multirow{2}{*}{2} & $k$UAPs & -&\textbf{	78.5}&70.6	&	0.85&	-&\textbf{	99.4}&	98.8&	10.16 \\ 
                 & &  T$_k$ML-AP-Uv  & -&	68&	59.5&	\textbf{0.16}&	-	&82&	82&	\textbf{0.15}  \\ \cline{2-11} 
&\multirow{2}{*}{3} & $k$UAPs &-	&-&	\textbf{79.5}&	2.09&	-&	-&	\textbf{73.2}&	10.16  \\ 
                 & &T$_k$ML-AP-Uv& -&	-&	65&	\textbf{0.17}&	-&	-&	80.5&	\textbf{0.16}\\ \hline
\multirow{6}{*}{0.8}&\multirow{2}{*}{1} & $k$UAPs&  \textbf{82.9}&	70.6&	60.7&0.66&	\textbf{96.7}&	78.6&	59.4&	2.60  \\ 
                  &&T$_k$ML-AP-Uv& 74.2&	60.7&	50.4&\textbf{0.14}&	84.5&	70.4&	63.7&	\textbf{0.13} \\ \cline{2-11} 
&\multirow{2}{*}{2} & $k$UAPs & -&	\textbf{84.2}&	77&0.98&	-&\textbf{99.3}&	98.1&1.62\\ 
                 & &  T$_k$ML-AP-Uv  & -&	70.5&	61.7&	\textbf{0.16}&	-	&81.2& 75.3&	\textbf{0.15}  \\ \cline{2-11} 
&\multirow{2}{*}{3} & $k$UAPs &-	&-&	$\times$&	$\times$&	-&	-&\textbf{98.6}&	3.95  \\ 
                 & &T$_k$ML-AP-Uv& -&	-&	\textbf{69}&	\textbf{0.18}&	-&	-&78.9&	\textbf{0.16}\\ \hline
\end{tabular}
\vspace{-1mm}
}
\caption{\small \it Comparison of \pert~and \asr~(\%) of the universal untargeted attack methods for two datasets. `$\times$' represents the current algorithm cannot get results. `-' represents the current results in the $k^{\prime}<k$ setting are the same as the results in the $k^{\prime}=k$ setting. $\epsilon=2000$. The best results are shown in bold.}
\label{tab:uap_ep_2000}
\vspace{-5mm}
\end{table}

In MS COCO 2014 dataset, the \asr~scores from our method are higher than the scores from the $k$UAPs. There is a huge gap (12.6\%) between two methods when $k=1$ and $\xi=0.7$. Second, we can find that the \pert~from our method is smaller than it from the $k$UAPs method in all $k$ value settings. On the other hand, we find the \asr~scores from our method are decreasing when increasing the $k$ value. However, there is no same trend in the $k$UAPs. The maximum score (74.6\%) can be achieved when $k=2$.  A potential explanation is that the $k$UAPs is based on the modified $k$Fool comparative method, which has no guarantee of the optimal solution in the optimization procedures. But our method is based on T$_k$ML-AP-U, which uses the sum of top-$k$ (a convex relaxation) in the optimization.  

\begin{table*}[th!]
% \captionsetup{font=footnotesize}
\centering
\renewcommand\arraystretch{0.5}
\setlength\tabcolsep{1.5pt}
\scriptsize{
\begin{tabular}{|c|c|c|cc|cc|cc|cc|cc|cc|}
\hline
\multirow{3}{*}{Cases} & \multirow{3}{*}{$k$} & \multirow{3}{*}{Methods} & \multicolumn{6}{c|}{PASCAL VOC 2012}                                                 & \multicolumn{6}{c|}{MS COCO 2014}                                                 \\ \cline{4-15} 
                  &                   &                   & \multicolumn{2}{c|}{$k^{\prime}$=3} & \multicolumn{2}{c|}{$k^{\prime}$=5} & \multicolumn{2}{c|}{$k^{\prime}$=10} & \multicolumn{2}{c|}{$k^{\prime}$=3} & \multicolumn{2}{c|}{$k^{\prime}$=5} & \multicolumn{2}{c|}{$k^{\prime}$=10} \\ \cline{4-15} 
                  & & & \pert($\times$10$^{-3}$)  & \asr & \pert($\times$10$^{-3}$)  & \asr & \pert($\times$10$^{-3}$) &\asr&\pert($\times$10$^{-3}$) &\asr &\pert($\times$10$^{-3}$)&\asr& \pert($\times$10$^{-3}$) &\asr   \\ \hline
\multirow{6}{*}{Best} & \multirow{2}{*}{3} &   ML-AP   & 3.04&	64.7&	2.63&	3.6&	1.14&	0.1&	4.96&	97.5&	4.65&	11.2&	4.35&	1.1 \\ 
                  & & T$_k$ML-AP-T   & 3.04&	\textbf{64.7}&	2.70&	2.80&	1.14&	0.1&	5.10&	97.5&	4.51&	9.1&	4.73&	0.7\\ \cline{2-15}
                  &\multirow{2}{*}{5}&    ML-AP     & -&-& 3.20&	51.5&	2.02&	0.5&	-&	-&	5.55&	94&	4.70&	3 \\ 
                  &&   T$_k$ML-AP-T     & -&- &3.20&	\textbf{51.8}&	1.96&	0.7&	-&	-&	5.68&	\textbf{94.5}&	5.14&	2.5 \\ \cline{2-15}
                  &\multirow{2}{*}{10}&  ML-AP  &  -&- & - &- & 3.33&	34.4&	-&	-&	-&	-&	6.10&	86.5\\ 
                  &&   T$_k$ML-AP-T   & -& - &-&- &3.37&	\textbf{35.5}&	-&	-	&-&	-&	6.22&	\textbf{87.8}\\ \cline{1-15}
\multirow{6}{*}{Random} &\multirow{2}{*}{3}& ML-AP &3.58&	34.3&	3.44&	12.9&	3.31&	1.6&	6.58&	84.5&	6.50&	45.2&	6.27&	14  \\ 
                  && T$_k$ML-AP-T  &3.60&	\textbf{38.2}&	3.46&	12.8&	3.49&	1.1&	6.63&	\textbf{86.3}&	6.55&	42.6&	6.52&	11.5\\ \cline{2-15}
                  &\multirow{2}{*}{5}& ML-AP  &-&-&3.67&	23.7&	3.82&	4.7&	-&	-&	6.72&	60.1&6.61&	22.9  \\
                  &&T$_k$ML-AP-T  & - & -&3.76&\textbf{	28.2}&	3.54&	3.7	&-&	-&6.81&\textbf{	68.1}&	6.65&	21.1 \\ \cline{2-15}
                  &\multirow{2}{*}{10}&ML-AP  &-&- & -&- & 3.75&	17.6&	-&	-&	-&	-	&6.95&	26.4  \\ 
                  && T$_k$ML-AP-T & -&-& -&-&3.80&	\textbf{20.5}&	-&	-&	-	&-&	6.97&\textbf{	42.6}  \\ \cline{1-15}
\multirow{6}{*}{Worst} &\multirow{2}{*}{3}& ML-AP  &3.88&	13.7&	3.85&	9&	3.82&	2.2&	6.71&	61.2&	6.68&	42.4&	6.78&	18.5  \\
                  &&  T$_k$ML-AP-T &3.85&	\textbf{17.5}&3.79&	10.2&3.74&	2.6&	6.76&	\textbf{65.5}&	6.74&	42&	6.75&	17.4 \\ \cline{2-15}
                  &\multirow{2}{*}{5}&  ML-AP &-&-&3.96&	8.3&	4.03&	3.4	&-&	-&	6.75&	39.9&6.70&	23.9\\
                  && T$_k$ML-AP-T & - & - &3.98&	\textbf{11.3}&	4.01&	3.3&	-&	-&	6.80&	\textbf{47.2}&	6.76&	23.2 \\ \cline{2-15}
                  &\multirow{2}{*}{10}&ML-AP & - & - & - &-&3.95&	6.2&	-&	-&	-&	-&6.90&	14.7  \\ 
                  &&T$_k$ML-AP-T  & - & -&- &-&4.04&	\textbf{10.2}&	-&	-&	-&	-&	6.88&	\textbf{24.4} \\ \hline
\end{tabular}
\vspace{-3mm}
\caption{\small \em Comparison of \pert~and \asr~(\%) of the targeted attack methods with $k$=3, 5, 10 in the Best, Random, and Worst cases on two datasets. The best \asr~results are shown in bold. $\epsilon=1$.}
\label{tab:targeted_ep_1}
}
\vspace{-5mm}
\end{table*}

When we compare the results between two datasets in our method, we can find that the \asr~scores from the MS COCO 2014 dataset are larger than ones from the PASCAL VOC 2012 dataset. The reason is that MS COCO 2014 has 80 labels. There are many labels (exclude the ground truth labels) that  can be pushed to the top-$k$ positions. However, there are only 20 labels in the PASCAL VOC 2012 dataset. Therefore, the attacking methods are easy to attack the baseline models with a dataset that contains more labels.

\begin{figure}[t]
\captionsetup[subfigure]{font=small}
\centering
\includegraphics[width=\linewidth]{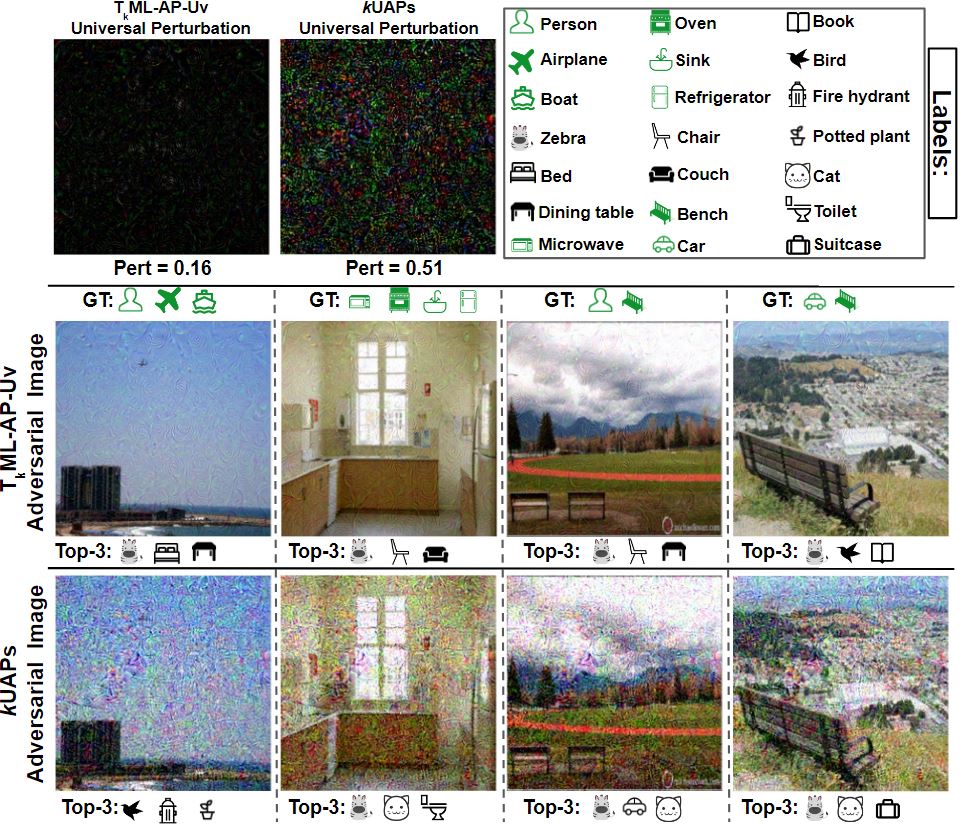}
\vspace{-5mm}
\caption{\small \it Examples of $k$UAPs and T$_k$ML-AP-Uv adversarial perturbations with $k=3$ on MS COCO 2014. GT means the ground truth labels. Top-3 means the Top-3 predicted labels from the corresponding image. The green icons correspond to the ground truth labels.}\label{fig:UUA_eps_100}
\vspace{-5mm}
\end{figure}

\begin{figure}[t]
\captionsetup[subfigure]{justification=centering, font=small}
\centering
\includegraphics[width=\linewidth]{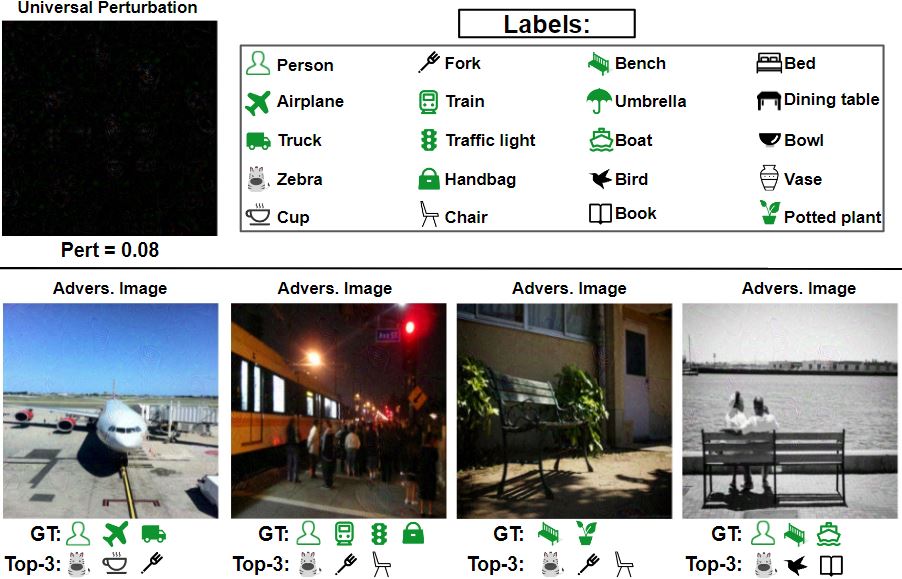}
\vspace{-5mm}
\caption{\small \it Examples of T$_k$ML-AP-Uv adversarial perturbations with $k=3$ on MS COCO 2014. GT means the ground truth labels. Top-3 means the Top-3 predicted labels from the corresponding image. Advers. means adversarial. The green icons correspond to the ground truth labels. $\epsilon=15$.}\label{fig:UUA_eps_15}
\vspace{-5mm}
\end{figure}

\begin{figure}[t]
\captionsetup[subfigure]{justification=centering, font=small}
\centering
\includegraphics[width=\linewidth]{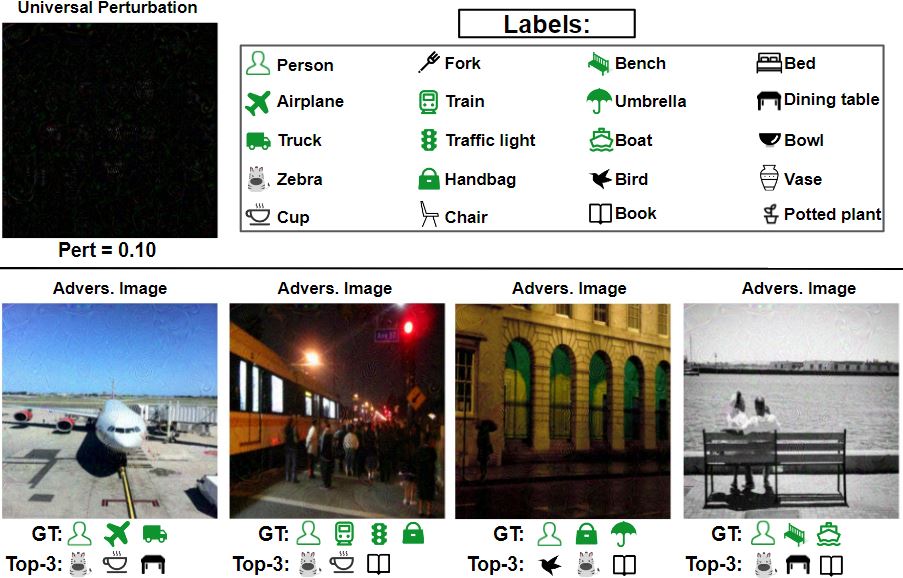}
\vspace{-5mm}
\caption{\small \it Examples of T$_k$ML-AP-Uv adversarial perturbations with $k=3$ on MS COCO 2014. GT means the ground truth labels. Top-3 means the Top-3 predicted labels from the corresponding image. Advers. means adversarial. The green icons correspond to the ground truth labels. $\epsilon=20$.}\label{fig:UUA_eps_20}
\vspace{-5mm}
\end{figure}

Since we set 100 as a projection threshold in the algorithm, the norm of $\z$ from the $k$UAPs method in all $k$ value settings can reach the maximum threshold, which means the perturbed images from $k$UAPs will distortion. When we increase the projection threshold, these values from the $k$UAPs will become larger. But the values from the T$_k$ML-AP-Uv method are stable. We show the results in Table \ref{tab:uap_ep_2000} with setting the projection threshold $\epsilon=2000$ .

\begin{table*}[th!]
% \captionsetup{font=footnotesize}
\centering
\renewcommand\arraystretch{0.5}
\setlength\tabcolsep{1.5pt}
\scriptsize{
\begin{tabular}{|c|c|c|cc|cc|cc|cc|cc|cc|}
\hline
\multirow{3}{*}{Cases} & \multirow{3}{*}{$k$} & \multirow{3}{*}{Methods} & \multicolumn{6}{c|}{PASCAL VOC 2012}                                                 & \multicolumn{6}{c|}{MS COCO 2014}                                                 \\ \cline{4-15} 
                  &                   &                   & \multicolumn{2}{c|}{$k^{\prime}$=3} & \multicolumn{2}{c|}{$k^{\prime}$=5} & \multicolumn{2}{c|}{$k^{\prime}$=10} & \multicolumn{2}{c|}{$k^{\prime}$=3} & \multicolumn{2}{c|}{$k^{\prime}$=5} & \multicolumn{2}{c|}{$k^{\prime}$=10} \\ \cline{4-15} 
                  & & & \pert($\times$10$^{-3}$)  & \asr & \pert($\times$10$^{-3}$)  & \asr & \pert($\times$10$^{-3}$) &\asr&\pert($\times$10$^{-3}$) &\asr &\pert($\times$10$^{-3}$)&\asr& \pert($\times$10$^{-3}$) &\asr   \\ \hline
\multirow{6}{*}{Best} & \multirow{2}{*}{3} &   ML-AP   & 4.44&96.2 &3.70& 15&2.42&0.2&5.49&100 &5.13&11.4&5.09&1.3 \\ 
                  & & T$_k$ML-AP-T   &4.45 &\textbf{96.6} &3.57&4.10 &1.14&0.1&5.71&\textbf{100} &5.55&11.4&5.68&0.9\\ \cline{2-15}
                  &\multirow{2}{*}{5}&    ML-AP     & -&-& 5.01 &92 &2.59& 0.6 &-&-&6.57 & 99.9&5.94 &3.9 \\ 
                  &&   T$_k$ML-AP-T     & -&- &5.02& \textbf{92.8} & 1.18&0.1&-&-&6.86&\textbf{99.9} &5.69 &2.9 \\ \cline{2-15}
                  &\multirow{2}{*}{10}&  ML-AP  &  -&- & - &- & 5.53& 84.2 &-&-&-&-&8.06&99.8\\ 
                  &&   T$_k$ML-AP-T   & -& - &-&- &5.59 & \textbf{86.4} &-&-&-&-&8.52&\textbf{99.8}\\ \cline{1-15}
\multirow{6}{*}{Random} &\multirow{2}{*}{3}& ML-AP &5.90&86&5.69&51.7& 4.88&3.1&9.43&99.8&9.35&60.7&9.17 &24.9  \\ 
                  && T$_k$ML-AP-T  &5.90 &\textbf{89.8} &5.39&25.3&4.57&2.4&9.89&\textbf{99.9}&9.79&58.4&9.72&22.7\\ \cline{2-15}
                  &\multirow{2}{*}{5}& ML-AP  &-&-&6.22	&77.9&	5.78&	16.3&	-&	-&	11.10&	96.5&	11.00&	43.2  \\
                  &&T$_k$ML-AP-T  & - & -&6.27&	\textbf{83.7}&	5.66&	10.5&	-&	-&	11.80&	\textbf{97.8}&	11.70&	42.7 \\ \cline{2-15}
                  &\multirow{2}{*}{10}&ML-AP  &-&- & -&- & 6.27	&67.7 &-& - &-&- &12.20&	84.2  \\ 
                  && T$_k$ML-AP-T & -&-& -&-&6.41&	\textbf{76.4} &-& -&-&-&12.80&	\textbf{94.5}  \\ \cline{1-15}
\multirow{6}{*}{Worst} &\multirow{2}{*}{3}& ML-AP  &6.59&	68&	6.43&	42.6&5.86&	7.7&	10.80&	90&	10.90&	72.2&	11.10&	43.3  \\
                  &&  T$_k$ML-AP-T &6.64&	\textbf{75.8}&	6.35&	36.2&	5.79&	7.3&	11.40&	\textbf{91.4}&	11.50&	71.6&	11.70&	43 \\ \cline{2-15}
                  &\multirow{2}{*}{5}&  ML-AP &-&-&6.75&53.3&	6.47&	18.8&	-&	-&	11.90&	81.8&	11.90&	56.6\\
                  && T$_k$ML-AP-T & - & - &6.90&\textbf{	66.6}&	6.41&	16.6&	-&	-&	12.50&	\textbf{87.2}&12.60&	58.9 \\ \cline{2-15}
                  &\multirow{2}{*}{10}&ML-AP & - & - & - &-&6.68&	39.1&	-&	-&	-&	-&	12.50&	59  \\ 
                  &&T$_k$ML-AP-T  & - & -&- &-&6.91&	\textbf{57}&	-&	-&	-&	-&	13.00&	\textbf{73.1} \\ \hline
\end{tabular}
\vspace{-3mm}
}
\caption{\small \it Comparison of \pert~and \asr~(\%) of the targeted attack methods with $k$=3, 5, 10 in the Best, Random, and Worst cases on two datasets. The best \asr~results are shown in bold. $\epsilon=2$.}
\label{tab:targeted_ep_2_old}
\vspace{-4mm}
\end{table*}

\begin{table*}[th!]
% \captionsetup{font=footnotesize}
\centering
\renewcommand\arraystretch{0.5}
\setlength\tabcolsep{1.5pt}
\scriptsize{
\begin{tabular}{|c|c|c|cc|cc|cc|cc|cc|cc|}
\hline
\multirow{3}{*}{Cases} & \multirow{3}{*}{$k$} & \multirow{3}{*}{Methods} & \multicolumn{6}{c|}{PASCAL VOC 2012}                                                 & \multicolumn{6}{c|}{MS COCO 2014}                                                 \\ \cline{4-15} 
                  &                   &                   & \multicolumn{2}{c|}{$k^{\prime}$=3} & \multicolumn{2}{c|}{$k^{\prime}$=5} & \multicolumn{2}{c|}{$k^{\prime}$=10} & \multicolumn{2}{c|}{$k^{\prime}$=3} & \multicolumn{2}{c|}{$k^{\prime}$=5} & \multicolumn{2}{c|}{$k^{\prime}$=10} \\ \cline{4-15} 
                  & & & \pert($\times$10$^{-3}$)  & \asr & \pert($\times$10$^{-3}$)  & \asr & \pert($\times$10$^{-3}$) &\asr&\pert($\times$10$^{-3}$) &\asr &\pert($\times$10$^{-3}$)&\asr& \pert($\times$10$^{-3}$) &\asr   \\ \hline
\multirow{6}{*}{Best} & \multirow{2}{*}{3} &   ML-AP   & 4.74&	98.9&	4.12&	5.6&	1.14&	0.1&	5.51&	100&	5.31&	12.2&	4.57&	1 \\ 
                  & & T$_k$ML-AP-T   & 4.76&	\textbf{99.2}&	3.91&	4.20&	1.14&	0.1&5.72&	\textbf{100}	&5.48&	11.1&4.91&	0.7\\ \cline{2-15}
                  &\multirow{2}{*}{5}&    ML-AP     & -&-& 5.53&	97.2&	5.29&	1.1&	-	&-&	6.60&	99.9&5.57&	3.3\\ 
                  &&   T$_k$ML-AP-T     & -&- &5.54&	\textbf{97.8}&	4.71&	1.1&	-&	-&	6.91&	\textbf{99.9}&	6.75&	3.9 \\ \cline{2-15}
                  &\multirow{2}{*}{10}&  ML-AP  &  -&- & - &- & 6.35&	94	&-	&-	&-&	-&	8.18&	100\\ 
                  &&   T$_k$ML-AP-T   & -& - &-&- &6.43&	\textbf{95.7}&	-&	-&	-&	-&	8.69&	\textbf{100}\\ \cline{1-15}
\multirow{6}{*}{Random} &\multirow{2}{*}{3}& ML-AP &6.77&	95.5&6.39&	40.2&	5.39&	2.8&	9.62&	100	&9.47&	64.6&	9.36&	25 \\ 
                  && T$_k$ML-AP-T  &6.72&	\textbf{97.7}&6.06&	29&5.03&	1.9&	10.10&	\textbf{100}	&10.00&	61.6&	9.94&	23.4\\ \cline{2-15}
                  &\multirow{2}{*}{5}& ML-AP  &-&-&7.32	&90.7&6.71&	17.2&	-&	-&	11.90&	98.5&	11.50&	45  \\
                  &&T$_k$ML-AP-T  & - & -&7.34&	\textbf{94.8}&	6.30&	12&	-&	-&	13.00&\textbf{	99.3}&12.90&	44.4 \\ \cline{2-15}
                  &\multirow{2}{*}{10}&ML-AP  &-&- & -&- & 7.55&	85&	-&	-&	-&	-	&14.40&	95  \\ 
                  && T$_k$ML-AP-T & -&-& -&-&7.74&\textbf{	91.8}&	-&	-&	-&	-&	16.50&	\textbf{99.1}  \\ \cline{1-15}
\multirow{6}{*}{Worst} &\multirow{2}{*}{3}& ML-AP  &7.92&	86.1&	7.75&	54.8&	7.09&	10.9&	11.90&	96.6&	12.10&	77.3&	12.60&	49.3  \\
                  &&  T$_k$ML-AP-T &8.04&	\textbf{92.9}&	7.67&	48.5&	6.83&	9.3&12.70&	\textbf{98}&	13.00&	77.7&13.50&	50.7 \\ \cline{2-15}
                  &\multirow{2}{*}{5}&  ML-AP &-&-&8.30&	75.7&7.67&	27.8&	-&	-&	13.50&	92.8&13.80&	67.8\\
                  && T$_k$ML-AP-T & - & - &8.58&	\textbf{89.2}&	7.74&	21.5&	-&	-&	15.00&	\textbf{95.8}&	15.30&	68.3 \\ \cline{2-15}
                  &\multirow{2}{*}{10}&ML-AP & - & - & - &-&8.37&	64.3&	-&	-&	-&	-&	15.50&	75.6  \\ 
                  &&T$_k$ML-AP-T  & - & -&- &-&8.84&	\textbf{83.6}&	-&	-	&-&	-&	17.90&\textbf{	90} \\ \hline
\end{tabular}
\vspace{-3mm}
\caption{\small \em Comparison of \pert~and \asr~(\%) of the targeted attack methods with $k$=3, 5, 10 in the Best, Random, and Worst cases on two datasets. The best \asr~results are shown in bold. $\epsilon=3$.}
\label{tab:targeted_ep_3}
}
\vspace{-4mm}
\end{table*}

\begin{table*}[th!]
% \captionsetup{font=footnotesize}
\centering
\renewcommand\arraystretch{0.5}
\setlength\tabcolsep{1.5pt}
\scriptsize{
\begin{tabular}{|c|c|c|cc|cc|cc|cc|cc|cc|}
\hline
\multirow{3}{*}{Cases} & \multirow{3}{*}{$k$} & \multirow{3}{*}{Methods} & \multicolumn{6}{c|}{PASCAL VOC 2012}                                                 & \multicolumn{6}{c|}{MS COCO 2014}                                                 \\ \cline{4-15} 
                  &                   &                   & \multicolumn{2}{c|}{$k^{\prime}$=3} & \multicolumn{2}{c|}{$k^{\prime}$=5} & \multicolumn{2}{c|}{$k^{\prime}$=10} & \multicolumn{2}{c|}{$k^{\prime}$=3} & \multicolumn{2}{c|}{$k^{\prime}$=5} & \multicolumn{2}{c|}{$k^{\prime}$=10} \\ \cline{4-15} 
                  & & & \pert($\times$10$^{-3}$)  & \asr & \pert($\times$10$^{-3}$)  & \asr & \pert($\times$10$^{-3}$) &\asr&\pert($\times$10$^{-3}$) &\asr &\pert($\times$10$^{-3}$)&\asr& \pert($\times$10$^{-3}$) &\asr   \\ \hline
\multirow{6}{*}{Best} & \multirow{2}{*}{3} &   ML-AP   & 4.83& 99.6&3.62& 4.7&1.14&0.1&5.50&100 &5.42&12.3&4.57&1.2 \\ 
                  & & T$_k$ML-AP-T   & 4.86&\textbf{99.8}&3.64 & 4.3&1.14&0.1&5.74&\textbf{100}&5.66&11.4&6.84&1.1\\ \cline{2-15}
                  &\multirow{2}{*}{5}&    ML-AP     & -&-& 5.68&98.4 &3.65& 1 &-&-&6.60 & 99.9&5.98 &3.8 \\ 
                  &&   T$_k$ML-AP-T     & -&- &5.73& \textbf{99.1} & 2.92&0.9&-&-&6.91&\textbf{99.9 }&5.96 &3.3 \\ \cline{2-15}
                  &\multirow{2}{*}{10}&  ML-AP  &  -&- & - &- & 6.62& 96.7 &-&-&-&-&8.19&100\\ 
                  &&   T$_k$ML-AP-T   & -& - &-&- &6.75 & \textbf{98.2} &-&-&-&-&8.70&\textbf{100}\\ \cline{1-15}
\multirow{6}{*}{Random} &\multirow{2}{*}{3}& ML-AP &7.00&97.4&6.66&40.6& 5.48&4&9.61&100&9.51&63.2&9.22 &24.6  \\ 
                  && T$_k$ML-AP-T  &6.96 &\textbf{99} &6.31&29.5&5.04&3.1&10.10&\textbf{100}&9.98&61.7&9.91&23.4\\ \cline{2-15}
                  &\multirow{2}{*}{5}& ML-AP  &-&-&7.61&93.6&6.64&16.3&-&-& 12.00&99&11.70&44.7  \\
                  &&T$_k$ML-AP-T  & - & -&7.84 &\textbf{98.6} &6.33&12.3&-&-&13.10& \textbf{99.7} & 13.10&44.6 \\ \cline{2-15}
                  &\multirow{2}{*}{10}&ML-AP  &-&- & -&- & 7.93 &89.1 &-& - &-&- &14.70& 96.8  \\ 
                  && T$_k$ML-AP-T & -&-& -&-&8.33 & \textbf{96.4} &-& -&-&-&17.20 &\textbf{99.9}  \\ \cline{1-15}
\multirow{6}{*}{Worst} &\multirow{2}{*}{3}& ML-AP  &8.35&90.9 &8.07&57.4 & 7.24&11.7 &12.10&97.9&12.30 &80.1 &13.10& 52.3  \\
                  &&  T$_k$ML-AP-T &8.66&\textbf{97.8} &8.09&50.6&7.09&10.9&12.90&\textbf{98.9} &13.30&77.4&13.90&50.9 \\ \cline{2-15}
                  &\multirow{2}{*}{5}&  ML-AP &-&-&8.75 &80.7&8.08 & 29.4 &-&- &13.80 &94.1 & 14.20 &68.9\\
                  && T$_k$ML-AP-T & - & - &9.47 &\textbf{96.2} &8.62 &25 &-&-&15.60&\textbf{98.3} &16.00 &72.2 \\ \cline{2-15}
                  &\multirow{2}{*}{10}&ML-AP & - & - & - &-&8.81&69.2&-&- &-&- &16.00&78.7  \\ 
                  &&T$_k$ML-AP-T  & - & -&- &-&10.10 &\textbf{94.7} &-&- &-&-& 19.90 & \textbf{95.7} \\ \hline
\end{tabular}
\vspace{-3mm}
\caption{\small \em Comparison of \pert~and \asr~(\%) of the targeted attack methods with $k$=3, 5, 10 in the Best, Random, and Worst cases on two datasets. The best \asr~results are shown in bold. $\epsilon=10$.}
\label{tab:targeted_ep_10}
}
\vspace{-5mm}
\end{table*}

\subsection{Additional Universal Untargeted Attacking Image results}\label{appendix:UAP_image_performance}

We set $\epsilon=100$ and $\xi=0.7$, then show the image results from both methods in Figure \ref{fig:UUA_eps_100}. However, we can set a smaller $\epsilon$ value and get a smaller perturbation. Here, we show more perturbed image results and report their top-3 predictions based on our T$_k$ML-AP-Uv method with setting different $\epsilon$. We set the projection threshold $\epsilon=15$ in Figure \ref{fig:UUA_eps_15} and $\epsilon=20$ in Figure \ref{fig:UUA_eps_20}.
% Additional image results in different threshold settings can be found in the Appendix \ref{appendix:UAP_image_performance}. With using the perturbation $\z=40$,  the perturbed images can fool the baseline models. This means our T$_k$ML-AP-UPs methods can success attack models in a universal level.

\subsection{Additional Targeted Attacking Performance}\label{appendix:TA_performance}
In the main paper, we set the projection threshold $\epsilon=2$ and report partial results. Here, we set more different projection thresholds such as $\epsilon=1$ in both T$_k$ML-AP-T and ML-AP algorithms and show the performance in the Table \ref{tab:targeted_ep_1}. We also show the complete results in Table \ref{tab:targeted_ep_2_old}, \ref{tab:targeted_ep_3} and \ref{tab:targeted_ep_10} with setting $\epsilon=2, 3, 10$, respectively. From these Tables, we can see that the performance is increasing when $\epsilon$ is increasing.

In the level of the cases, we find that the \asr~score is decreasing when the selected labels are hard to attack in the same $k$ values. For example, for $k^{\prime}=3$ in the PASCAL VOC 2014 dataset with $\epsilon=2$, the \asr~score of the ML-AP method is decreased from 96.2\% (Best) to 86\% (Random), which has a 10.2\% difference. This difference becomes large from 86\% (Random) to 68\% (Worst), which has an 18\% difference. On the other hand, for our method, we can find the difference is 6.8\% from 96.6\% (Best) to 89.8\% (Random), and the difference becomes 14\% from the 89.8\% (Random) to 75.8\% (Worst). Comparing these difference values between the two methods in the same scenario, we can see that the difference from T$_k$ML-AP-T is smaller than the difference of ML-AP. This means our method is more robust than the ML-AP method. For the \pert~values of perturbation norms, we find that it is increased when the selected labels are hard to attack in the same $k$ values. This is very intuitive because both methods need to add more perturbations to handle the hard tasks.

%% file: main.bbl
\begin{thebibliography}{10}\itemsep=-1pt

\bibitem{akhtar2018threat}
Naveed Akhtar and Ajmal Mian.
\newblock Threat of adversarial attacks on deep learning in computer vision: A
  survey.
\newblock {\em IEEE Access}, 6:14410--14430, 2018.

\bibitem{boyd2004convex}
Stephen Boyd, Stephen~P Boyd, and Lieven Vandenberghe.
\newblock {\em Convex optimization}.
\newblock Cambridge university press, 2004.

\bibitem{carlini2017towards}
Nicholas Carlini and David Wagner.
\newblock Towards evaluating the robustness of neural networks.
\newblock In {\em 2017 ieee symposium on security and privacy (sp)}, pages
  39--57. IEEE, 2017.

\bibitem{everingham2015pascal}
Mark Everingham, SM~Ali Eslami, Luc Van~Gool, Christopher~KI Williams, John
  Winn, and Andrew Zisserman.
\newblock The pascal visual object classes challenge: A retrospective.
\newblock {\em International journal of computer vision}, 111(1):98--136, 2015.

\bibitem{fan2017learning}
Yanbo Fan, Siwei Lyu, Yiming Ying, and Baogang Hu.
\newblock Learning with average top-k loss.
\newblock In {\em Advances in neural information processing systems}, pages
  497--505, 2017.

\bibitem{goodfellow2014explaining}
Ian~J Goodfellow, Jonathon Shlens, and Christian Szegedy.
\newblock Explaining and harnessing adversarial examples.
\newblock {\em International Conference on Learning Representations}, 2015.

\bibitem{he2016deep}
Kaiming He, Xiangyu Zhang, Shaoqing Ren, and Jian Sun.
\newblock Deep residual learning for image recognition.
\newblock In {\em Proceedings of the IEEE conference on computer vision and
  pattern recognition}, pages 770--778, 2016.

\bibitem{hu2020learning}
Shu Hu, Yiming Ying, Siwei Lyu, et~al.
\newblock Learning by minimizing the sum of ranked range.
\newblock {\em Advances in Neural Information Processing Systems}, 33, 2020.

\bibitem{kurakin2016adversarial}
Alexey Kurakin, Ian Goodfellow, and Samy Bengio.
\newblock Adversarial machine learning at scale.
\newblock {\em International Conference on Learning Representations}, 2017.

\bibitem{lapin2015top}
Maksim Lapin, Matthias Hein, and Bernt Schiele.
\newblock Top-k multiclass svm.
\newblock {\em Advances in neural information processing systems}, 2015.

\bibitem{lin2014microsoft}
Tsung-Yi Lin, Michael Maire, Serge Belongie, James Hays, Pietro Perona, Deva
  Ramanan, Piotr Doll{\'a}r, and C~Lawrence Zitnick.
\newblock Microsoft coco: Common objects in context.
\newblock In {\em European conference on computer vision}, pages 740--755.
  Springer, 2014.

\bibitem{liu2016delving}
Yanpei Liu, Xinyun Chen, Chang Liu, and Dawn Song.
\newblock Delving into transferable adversarial examples and black-box attacks.
\newblock {\em International Conference on Learning Representations}, 2017.

\bibitem{lyu2018univariate}
Siwei Lyu and Yiming Ying.
\newblock A univariate bound of area under roc.
\newblock In {\em Proceedings of the Conference on Uncertainty on Artificial
  Intelligence (UAI)}, 2018.

\bibitem{madry2017towards}
Aleksander Madry, Aleksandar Makelov, Ludwig Schmidt, Dimitris Tsipras, and
  Adrian Vladu.
\newblock Towards deep learning models resistant to adversarial attacks.
\newblock {\em International Conference on Learning Representations}, 2018.

\bibitem{melacci2020can}
Stefano Melacci, Gabriele Ciravegna, Angelo Sotgiu, Ambra Demontis, Battista
  Biggio, Marco Gori, and Fabio Roli.
\newblock Can domain knowledge alleviate adversarial attacks in multi-label
  classifiers?
\newblock {\em arXiv preprint arXiv:2006.03833}, 2020.

\bibitem{moosavi2017universal}
Seyed-Mohsen Moosavi-Dezfooli, Alhussein Fawzi, Omar Fawzi, and Pascal
  Frossard.
\newblock Universal adversarial perturbations.
\newblock In {\em Proceedings of the IEEE conference on computer vision and
  pattern recognition}, pages 1765--1773, 2017.

\bibitem{moosavi2016deepfool}
Seyed-Mohsen Moosavi-Dezfooli, Alhussein Fawzi, and Pascal Frossard.
\newblock Deepfool: a simple and accurate method to fool deep neural networks.
\newblock In {\em Proceedings of the IEEE conference on computer vision and
  pattern recognition}, pages 2574--2582, 2016.

\bibitem{ogryczak2003minimizing}
Wlodzimierz Ogryczak and Arie Tamir.
\newblock Minimizing the sum of the k largest functions in linear time.
\newblock {\em Information Processing Letters}, 85(3):117--122, 2003.

\bibitem{papernot2016limitations}
Nicolas Papernot, Patrick McDaniel, Somesh Jha, Matt Fredrikson, Z~Berkay
  Celik, and Ananthram Swami.
\newblock The limitations of deep learning in adversarial settings.
\newblock In {\em 2016 IEEE European symposium on security and privacy
  (EuroS\&P)}, pages 372--387. IEEE, 2016.

\bibitem{ridnik2021tresnet}
Tal Ridnik, Hussam Lawen, Asaf Noy, Emanuel Ben~Baruch, Gilad Sharir, and
  Itamar Friedman.
\newblock Tresnet: High performance gpu-dedicated architecture.
\newblock In {\em Proceedings of the IEEE/CVF Winter Conference on Applications
  of Computer Vision}, pages 1400--1409, 2021.

\bibitem{russakovsky2015imagenet}
Olga Russakovsky, Jia Deng, Hao Su, Jonathan Krause, Sanjeev Satheesh, Sean Ma,
  Zhiheng Huang, Andrej Karpathy, Aditya Khosla, Michael Bernstein, et~al.
\newblock Imagenet large scale visual recognition challenge.
\newblock {\em International journal of computer vision}, 115(3):211--252,
  2015.

\bibitem{song2018multi}
Qingquan Song, Haifeng Jin, Xiao Huang, and Xia Hu.
\newblock Multi-label adversarial perturbations.
\newblock In {\em 2018 IEEE International Conference on Data Mining (ICDM)},
  pages 1242--1247. IEEE, 2018.

\bibitem{szegedy2016rethinking}
Christian Szegedy, Vincent Vanhoucke, Sergey Ioffe, Jon Shlens, and Zbigniew
  Wojna.
\newblock Rethinking the inception architecture for computer vision.
\newblock In {\em Proceedings of the IEEE conference on computer vision and
  pattern recognition}, pages 2818--2826, 2016.

\bibitem{szegedy2013intriguing}
Christian Szegedy, Wojciech Zaremba, Ilya Sutskever, Joan Bruna, Dumitru Erhan,
  Ian Goodfellow, and Rob Fergus.
\newblock Intriguing properties of neural networks.
\newblock {\em International Conference on Learning Representations}, 2014.

\bibitem{tursynbek2020geometry}
Nurislam Tursynbek, Aleksandr Petiushko, and Ivan Oseledets.
\newblock Geometry-inspired top-k adversarial perturbations.
\newblock {\em arXiv preprint arXiv:2006.15669}, 2020.

\bibitem{zhang2020learning}
Zekun Zhang and Tianfu Wu.
\newblock Learning ordered top-k adversarial attacks via adversarial
  distillation.
\newblock In {\em Proceedings of the IEEE/CVF Conference on Computer Vision and
  Pattern Recognition Workshops}, pages 776--777, 2020.

\bibitem{zhou2020generating}
Nan Zhou, Wenjian Luo, Xin Lin, Peilan Xu, and Zhenya Zhang.
\newblock Generating multi-label adversarial examples by linear programming.
\newblock In {\em 2020 International Joint Conference on Neural Networks
  (IJCNN)}, pages 1--8. IEEE, 2020.

\end{thebibliography}
